%% file: main.tex
\documentclass[manuscript,screen, acmsmall]{acmart}

\usepackage{graphicx}
\usepackage{subcaption}
\usepackage{float}
\usepackage{titlesec}
\usepackage{ragged2e}
\usepackage{amsmath}
\usepackage{algorithm}
\usepackage{algorithmic}
\usepackage{overpic}
\usepackage{xcolor}  

\AtBeginDocument{%
  }

\setcopyright{acmlicensed}
\copyrightyear{2025}
\acmYear{2025}
\acmDOI{XXXXXXX.XXXXXXX}
\begin{document}

\title{Self-supervised New Activity Detection
in Sensor-based Smart Environments}

\author{Hyunju Kim}
\email{iplay93@kaist.ac.kr}
\orcid{1234-5678-9012}
\affiliation{%
  \institution{Korea Advanced Institute of Science \& Technology}
  \city{Daejeon}
  \country{South Korea}
}

\author{Dongman Lee}
\email{dlee@kaist.ac.kr}
\orcid{1234-5678-9012}
\affiliation{%
  \institution{Korea Advanced Institute of Science \& Technology}
  \city{Daejeon}
  \country{South Korea}}

\renewcommand{\shortauthors}{Kim et al.}

\begin{abstract}
With the rapid advancement of ubiquitous computing technology, human activity analysis based on time series data from a diverse range of sensors enables the delivery of more intelligent services. 
Despite the importance of exploring new activities in real-world scenarios, existing human activity recognition studies generally rely on predefined known activities and often overlook detecting \textit{new patterns (novelties) that have not been previously observed during training}.
Novelty detection in human activities becomes even more challenging due to (1) diversity of patterns within the same known activity, (2) shared patterns between known and new activities, and (3) differences in sensor properties of each activity dataset.
We introduce \texttt{CLAN}, a two-tower model 
that leverages \textbf{C}ontrastive \textbf{L}earning with diverse data \textbf{A}ugmentation for \textbf{N}ew activity detection in sensor-based environments.
\texttt{CLAN} simultaneously and explicitly utilizes multiple types of strongly shifted data as negative samples in contrastive learning, effectively learning invariant representations that adapt to various pattern variations within the same activity. 
To enhance the ability to distinguish between known and new activities that share common features, \texttt{CLAN} incorporates both time and frequency domains, enabling the learning of multi-faceted discriminative representations.
Additionally, we design an automatic selection mechanism of data augmentation methods tailored to each dataset’s properties, generating appropriate positive and negative pairs for contrastive learning. 
Comprehensive experiments on real-world datasets show that \texttt{CLAN} achieves a 9.24\% improvement in AUROC compared to the best-performing baseline model.
\end{abstract}




\maketitle

\input{content/1.introduction}
\input{content/2.relatedwork}

\input{content/3.method}
\input{content/4.setup}

\input{content/5.results}
\input{content/6.discussion}

\input{content/7.conclusion}

\begin{acks}

This work was supported by the Institute for Information \& communication Technology Planning \& evaluation(IITP) funded by the Korea government(MSIT) (No.II191126, Self-learning based Autonomic IoT Edge Computing \& No.00459749, AI based Multiplex Smart Drug Detection Solution Development in Contactless Manner) and the National Research Foundation of Korea(NRF) grant funded by the Korean government(MSIT) (No.NRF-2022M3J6A1063021, Industry and Society Demand Oriented Open Human Resource Development).
\end{acks}

\bibliographystyle{ACM-Reference-Format}
\bibliography{ref}

\appendix
\input{content/appendix}








\end{document}

%% file: content/1.introduction.tex
\section{Introduction} \label{intro}

The advancement of ubiquitous technology enables surrounding devices \cite{WoT} to provide intelligent services that enhance user convenience and efficiency across various domains, including ambient assisted living, healthcare, and smart automation systems \cite{application_1, application_3, proactive}.
Human Activity Recognition (HAR) plays a crucial role in understanding user behaviors by analyzing sensor data collected from the smart devices \cite{HAR_survey_3, data_shift, TF_wearable}. 
The American Time Use Survey \cite{american_survey, Nuactiv} identifies over 462 distinct daily activities, which further vary depending on environmental factors and personal preferences \cite{HAR_survey_4}.
Given the vast number of potential real-world activities, defining every possible activity is impractical, leading to the continuous emergence of \textit{previously unseen activities}. 
Ignoring these new activities may reduce comfort in personalized automation services \cite{application_3, HAR_survey_1} or hinder timely responses to unforeseen behavioral changes in healthcare systems \cite{application_1}. 
This highlights the need for novelty detection techniques to identify new activities, enabling the development of adaptable intelligent systems in \textit{open-world environments} \cite{lifespan, novelty_sv2, discovery}.

Existing novelty detection methods \cite{anomaly3, DCdetector, ssl_har_sv, ssl_har_sv_2} in HAR primarily assume that new activities share some attributes or data with those seen during training. 
They \textit{cannot be applied to detect entirely new activities that were absent during training} (issue 1).
Moreover, they focus on detecting novelties at the level of individual data points or short subsequences.
Real-world human activities are inherently complex as they involve multiple sensors, user interactions, and fundamental temporal dynamics \cite{dynamics, group_dynamics}.
Due to this complexity, the existing detection methods may perform well for certain sensor types (e.g., IMUs) but often misclassify natural variations within known activities as novelties.
This misclassification can lead to severe performance degradation \textit{in diverse types of sensor environments}, rendering these methods impractical for new activity detection in real-world scenarios (issue 2).

Inspired by advances in other domains,  \textit{instance-level self-supervised learning (SSL) based novelty detection} \cite{novelty_sv4, novelty_sv1} offers a potential solution to the issues above.
Specifically, approaches leveraging contrastive learning have demonstrated strong capabilities in identifying new patterns through instance discrimination by analyzing \textit{entire data sequences} rather than individual data points or short subsequences, \textit{using only known data} \cite{ILD, novelty_sv1, novelty_sv3, FewSome}. 
They learn representations by pulling similar samples (positives) closer together while pushing dissimilar samples (negatives) apart, enabling the model to capture the key characteristics of known classes effectively.
Since they do not rely on data from new classes, they typically employ data augmentation techniques to generate positive and negative samples from known class instances \cite{novelty_cv2, DE_CL, SimCLR, GOAD}. 
However, despite their general applicability, applying existing contrastive learning-based approaches to HAR remains challenging due to its unique complexities. 
These challenges, as illustrated in Fig. \ref{fig:problem}, are as follows:
\begin{figure}[t]
    \centering
    \subfloat[Pattern diversity within the same known activity.]{
        \includegraphics[width=0.34\textwidth]{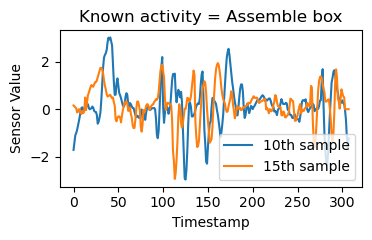}}
    \hspace{1mm}
    \subfloat[Similar patterns between known and new activities.]{
        \includegraphics[width=0.32\textwidth]{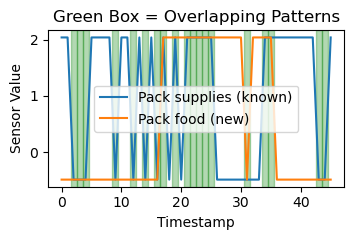}
    }        
    \hfill
    \subfloat[Different effects of \textit{Permute} data augmentation depending on dataset properties (Top: \textit{DOO-RE}, Bottom: \textit{OPENPACK})]{
    \includegraphics[width=0.68\textwidth]{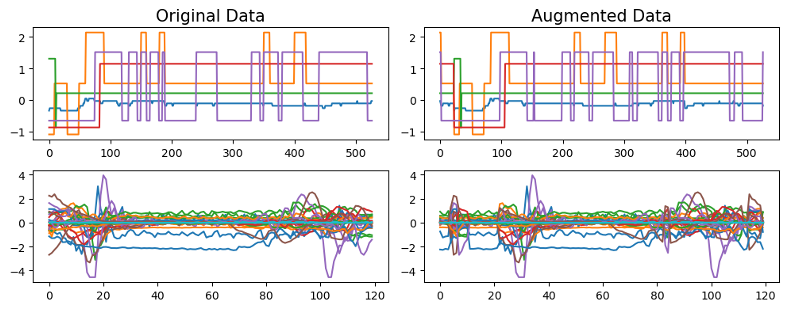}}
    \caption{Challenges faced by the existing novelty detection methods when applied to human activities.
    }
    \label{fig:problem}   

\end{figure}

(a)
\textit{Diverse pattern variations within the same known activity:}
Human behavior patterns exhibit diversity for the same activity due to temporal dynamics, inherent behavior dynamics, noise, or irregular sampling intervals \cite{dynamics, group_dynamics, data_shift, TFC}.
If a single type of highly shifting data augmentation method (e.g., rotation in computer vision) is applied to generate negatives, as done in previous methods, the model tends to learn overfitted representations for that specific variation.
This hinders the model's ability to generalize to other types of variations, thereby reducing its overall robustness in the real world.

(b) 
\textit{Overlapping temporal patterns between known and new activities:}  
Even when activities differ, humans often share fundamental movements or actions across various activities, leading to overlapping temporal patterns between known and new activities \cite{complex_activity}.  
When representations focus on a single aspect for detecting new classes, as shown in previous studies, similar patterns tend to cluster densely within the constrained feature space of the human activity domain, making activity differentiation more challenging.

(c) \textit{Differences in properties for each dataset:}
Human activity datasets exhibit varying properties, such as differences in sensor modalities, value ranges, and activity durations, which influence the effectiveness of data augmentation techniques \cite{TF_wearable}.
Applying uniform augmentation methods across all datasets, as is common in prior research, often leads to suboptimal negative sample generation.
This blurs the decision boundaries between known and new activities, resulting in degraded representation learning and impairing the model’s ability to general use in diverse types of sensor environments.

In this paper, we propose \texttt{CLAN},  a two-tower model that leverages \textbf{C}ontrastive \textbf{L}earning with diverse data \textbf{A}ugmentation for \textbf{N}ew activity detection in sensor-based environments.
\texttt{CLAN} aims to learn discriminative representations for known activities \textit{without accessing data from new activities} and leverages these representations to detect new activity instances.
(1) \texttt{CLAN} learns invariant representations that are robust to pattern diversity by explicitly and simultaneously pushing various types of strongly shifted samples as negatives away from the original known activity data in contrastive learning.
(2) To learn discriminative representations even when similar temporal patterns occur between known and new activities, \texttt{CLAN} is designed with a two-tower structure that decomposes data into time and frequency domains.
(3) \texttt{CLAN} selects the appropriate data augmentation methods for generating negatives tailored to each human activity dataset by performing a classification task between original and augmented samples.
To meet the design requirements, \texttt{CLAN} operates through three stages, as illustrated in Fig. \ref{fig3: overview}: customized strong transformation set construction, discriminative representation learning, and new activity detection.
The main \textbf{contributions} of our approach are summarized as follows: 
\begin{itemize} 
\item We propose \texttt{CLAN}, a two-tower self-supervised new activity detection model that extracts multi-faceted discriminative representations using contrastive learning, explicitly comparing multiple types of negatives in both the time and frequency domains. 
\item We design a classification task that enables the automatic selection of data augmentation methods, ensuring effective negative sample generation tailored to the properties of each human activity dataset.
\item Quantitative and qualitative evaluations of \texttt{CLAN} against representative novelty detection baselines across five different types of human activity datasets demonstrate that \texttt{CLAN} consistently outperforms existing methods, achieving an average improvement of 9.24\% in AUROC and 14.66\% in Balanced Accuracy.
\end{itemize}

%% file: content/2.relatedwork.tex
\section{Related Work} \label{relatedwork}
In this section, we explain research that we draw inspiration from while developing \texttt{CLAN} and discuss their constraints in detecting novel patterns in human activity data.

\subsection{Human Activity Recognition in Sensor-based Smart Environments}
Human Activity Recognition (HAR) using sensors in smart environments \cite{HAR_survey_1, HAR_survey_2} has become a fundamental technology for delivering optimal user services with minimal disruption \cite{aal_1, aal_2}.
Recent advancements have introduced methods that leverage or hybridize various deep learning techniques \cite{HAR_survey_3, HAR_survey_4, deep_sv}, enabling more comprehensive and sophisticated modeling of human activity data.
Many recent studies \cite{THAT, TST, ART} have explored Transformer variants for HAR, with their multi-head attention mechanisms effectively capturing temporal dependencies.
Additionally, recent research \cite{TFC, TFAD, HAR_survey_7} has emphasized the importance of sensor frequency information in representing time series data, extending beyond temporal dependency modeling.
Both time and frequency domains are inherent in all sensor-driven time series data \cite{TS-analysis}, and leveraging them together improves the comprehensive analysis of human activity datasets.
Analyzing the time domain provides insights into statistical characteristics such as the mean and variance of individual sensor values, as well as temporal correlations between different sensors.
Exploring the frequency domain reveals patterns in sensor activation frequencies and the overall distribution of sensor occurrences within each activity instance.

Although previous approaches for HAR have demonstrated notable success, they struggle to adapt and remain flexible when encountering new activities in dynamic and evolving real-world environments \cite{american_survey, Nuactiv, HAR_survey_4}.

\subsection{Novelty Detection}
Novelty detection, widely used across various applications, involves identifying unique or previously unseen patterns in data that deviate from the training distribution \cite{novelty_sv1, novelty_sv2, novelty_sv3}.
In sensor-based time series analysis, traditional novelty detection methods primarily focus on identifying novelties at the level of individual data points or short subsequences \cite{novelty_sv4, anomaly2, anomaly3}, particularly in datasets with repeating patterns.
The existing novelty detection methods assume that new activities share certain attributes or data patterns with those seen during training, making them ineffective for detecting entirely new activities that were absent from the training phase.
Moreover, the complexity of the human activity, with its intricate temporal dynamics \cite{dynamics}, often leads to frequent false novelty alarms, making it difficult to accurately identify genuinely new patterns at these levels.

In computer vision and natural language processing, cutting-edge research has introduced a diverse range of sophisticated techniques for instance-level novelty detection \cite{novelty_sv1, novelty_sv2, nlp_sv}.
A common mechanism shared by prior approaches involves:
1) leveraging encoding and model learning techniques to extract invariant representations of known classes, and
2) estimating novelty detection scores that differentiate between known and new classes.

Reconstruction-based \cite{novelty_cv1, reconstruction_based, DE_CL, ocgan} and classification-based \cite{novelty_nlp, one_class, ocsvm} methods are particularly prominent.
Reconstruction-based techniques commonly employ encoder-decoder frameworks that are trained with in-distribution data, utilizing models like autoencoders (AEs) \cite{autoencoder,reconstruction_based} or Generative Adversarial Networks (GANs) \cite{GAN,ocgan}.
One example is OCGAN \cite{ocgan}, which uses a denoising auto-encoder network for one-class novelty detection. 
This approach constrains a latent space to exclusively represent the given known class by employing bounded support, adversarial training in the latent space, and gradient-descent-based sampling.
Classification-based methods, on the other hand, use the output of a neural network classifier as a novelty detection score to determine whether incoming data belongs to a new class.
Ruff \textit{et al.} introduce \textit{DeepSVDD} \cite{one_class}, a kernel-based one-class classification approach that trains deep neural networks with a minimum volume estimation objective, forming a compact and representative hypersphere boundary for the known class.
Recently, self-supervised learning research \cite{SimCLR, novelty_cv2, GOAD, FewSome, UNODE}, particularly in contrastive learning, has shown significant advancements in detecting novel patterns.

\subsection{Contrastive Learning} \label{RW_CL}
Contrastive learning \cite{SimCLR, ILD} commonly extracts invariant representations by maximizing the similarity between representations of positive pairs while minimizing the similarity between negative pairs.
Recent studies have explored the application of contrastive learning to HAR \cite{ssl_har_sv, ssl_har_sv_2, TFC}; however, these methods primarily focus on classifying predefined activities, making them challenging to apply to new activity detection tasks directly.
In other domains, contrastive learning \cite{novelty_cv2, GOAD, FewSome, UNODE} has been leveraged for novelty detection by training models to extract invariant representations of known classes and identifying patterns that deviate from them.
A key challenge in novelty detection is the inability to access information about new classes during training, prompting the development of methods that improve representations through augmented samples.
For example, Tack \textit{et al.} \cite{novelty_cv2} introduced the CSI framework, which enhances inlier representations by treating strongly shifted samples (e.g., rotated images) as negatives in contrastive learning.

Despite its potential, the existing novelty detection techniques struggle to effectively capture key features relevant to human activity data in diverse types of sensor-based smart environments.

%% file: content/3.method.tex
\section{Method}
\begin{figure*}[h]
    \centering
    \begin{overpic}[width=\textwidth]{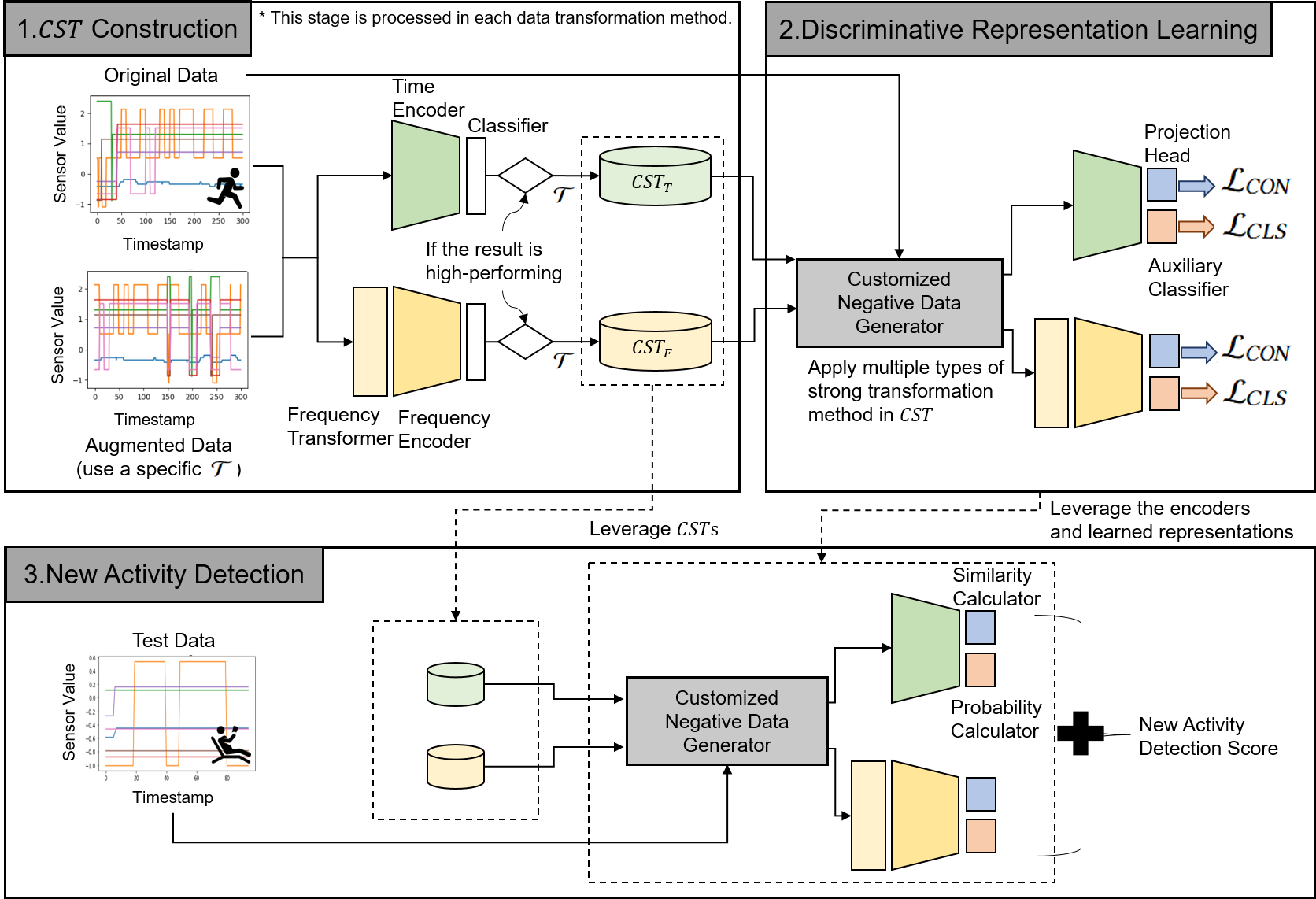}
    \end{overpic}
    
    \caption{Illustration of \texttt{CLAN} for new activity detection in sensor-based smart environments.}
    \label{fig3: overview}
\end{figure*}

In this section, we define the problem statement for detecting new human activities and provide an in-depth overview of \texttt{CLAN}'s structure and its detailed modules, as illustrated in Fig. \ref{fig3: overview}.

\subsection{Problem Definition: New Activity Detection } \label{preliminaries} \label{def2}
Given a training human activity dataset $\mathcal{X}=\{x_{i}\}_{i =1}^{N}$ containing $N$ samples, each sample $x_{i}$ represents an activity episode consisting of sensor data \cite{HAR_survey_1,changepoint}.  
Each $x_{i} \in \mathbb{R}^{D \times L}$ consists of $D$ sensor dimensions across $L$ timestamps, where $D=1$ for univariate data and $D>1$ for multivariate data.  
The distribution of $\mathcal{X}$, denoted as $p(x)$, defines the in-distribution, meaning that any sample drawn from $p(x)$ corresponds to a known activity.

The objective of \texttt{CLAN} is to train the \textit{Encoder} $g_{\Theta}$ to extract representations from $p(x)$ that are:  
(a) invariant to variations within the same activity,  
(b) discriminative for overlapping patterns between known and new activities, and  
(c) tailored to the properties of $\mathcal{X}$.  
Importantly, the training domain $X_{s}$ and the inference domain $X_{t}$ share the same feature space, meaning $X_{s} = X_{t}$, but their label spaces differ, such that $Y_{s} \neq Y_{t}$.  
The \textit{New Activity Detection Score} $sc_{\texttt{CLAN}}$ determines whether $x_{\text{test}}$ belongs to $p(x)$ using the representations extracted by $g_{\Theta}$.

\subsection{Overall Structure}

The two-tower model \texttt{CLAN} (\textbf{C}ontrastive \textbf{L}earning with diverse data \textbf{A}ugmentation for \textbf{N}ew activity detection) is proposed, consisting of three key stages (Fig.~\ref{fig3: overview}):  
(1) The \textit{CST Construction} stage involves a classification task to automatically decide the Customized Strong Transformation set (\textit{CST}), i.e., data augmentation methods that generate strongly shifted samples tailored to each dataset. 
This stage ensures the generation of appropriate positive and negative samples.  
(2) In the \textit{Discriminative Representation Learning} stage, \texttt{CLAN} extracts invariant representations of known activities through contrastive learning and auxiliary classification, leveraging original and diverse negative samples generated using \textit{CST}.  
To further differentiate overlapping features of known and new activities, invariant representation learning is applied separately to the time and frequency domains, deriving discriminative representations from a multi-faceted perspective.  
This process (a) consolidates key patterns of known activities while filtering out meaningless variations and (b) expands representations of known activities from multiple perspectives within the semantic scope \cite{data_shift, DE_CL}.  
(3) The \textit{New Activity Detection} stage identifies new activity instances by computing new activity detection scores based on similarity measures and probability estimates using the learned representations from the \textit{Discriminative Representation Learning} stage.

\subsection{Customized Strong Transformation Set (CST) Construction} \label{construction}
 
The data augmentation methods and data-shifting techniques used in previous studies \cite{SimCLR, novelty_cv2, DE_CL} are not well-suited for \textit{human activity data in diverse sensor environments}.  
Applying the same data transformations across all datasets for generating positive or negative samples can hinder the learning of precise representation boundaries for known activities, as dataset-specific sensor properties vary.  
To address this, we design an instance-shifting mechanism that determines the degree of transformation for various data augmentation methods, ensuring their suitability for generating positive or negative samples in each dataset.

Quantitatively assessing the degree of shifting is achieved by measuring the dissimilarity between original and augmented samples, formulated as a classification task with binary cross-entropy (BCE) loss \cite{dissimilarity}.  
The classification model $f_{\text{aug\_cls}}: \mathcal{X}_{\text{total}} \rightarrow \{0, 1\}$, where $\mathcal{X}_{\text{total}} = \mathcal{X} \cup \mathcal{X}_{\text{aug}}$, follows the objective function:
\begin{equation} \label{dg}
\mathcal{L}_{\text{aug\_cls}} = - \frac{1}{|\mathcal{X}_{\text{total}}|} \sum_{i=1}^{|\mathcal{X}_{\text{total}}|} \left( y_i \log(\hat{y}_i) + (1 - y_i) \log(1 - \hat{y}_i) \right)
\end{equation}
where $y_i$ is the true label for $x_i$, assigned as 0 if $x_i \in \mathcal{X}$ and 1 if $x_i \in \mathcal{X}_{\text{aug}}$.  
$\hat{y}_i$ represents the predicted probability of $x_i$, obtained using a sigmoid function.  
The model encoding architecture is identical to the encoder used in the \textit{Discriminative Representation Learning} stage for consistency.

For each data transformation method \(\mathcal{T}\), we evaluate the classification performance \(AUROC_{\mathcal{T}}\) using \(f_{\text{aug\_cls}}\), where a high \(AUROC_{\mathcal{T}}\) value indicates that \(\mathcal{T}\) induces a significant shift, making it suitable for negative sample generation.  
Data augmentation methods with \(AUROC_{\mathcal{T}}\) values exceeding a predefined threshold \(\theta_{CST}\) are included in the \textit{Customized Strong Transformation} (\(CST\)) set, defined as \(CST = \{\mathcal{T}_{i}\}_{i=1}^{K}\), where \(K\) represents the number of selected strong transformation types.

The \textit{Customized Negative Data Generator} utilizes \textit{CST} to generate diverse negative samples for subsequent stages by applying the \( j \)-th data transformation method from \textit{CST} to \( x \), denoted as \( x_{(j)}\).  
To minimize overlap with negative samples in the data space, the transformation method with the lowest \(AUROC_{\mathcal{T}}\), which is ensured to be weakly shifted from the original data, is selected to generate positive samples as \(\Tilde{x}\).

\subsection{Discriminative Representation Learning} \label{encoder}

\subsubsection{General Contrastive Learning.} \label{theoretical}

Drawing inspiration from instance-level novelty detection in the computer vision domain \cite{novelty_cv2, DE_CL, SimCLR, GOAD}, the general goal of contrastive learning is based on the principle of instance discrimination \cite{ILD}, where an \textit{Encoder} $g_{\Theta}$ is trained by maximizing the similarity between positive pairs while minimizing it between negative pairs.  
We leverage the NT-Xent contrastive loss \cite{SimCLR, novelty_cv2, TF_wearable}:
\begin{equation}\label{contrastiveloss}
\mathcal{L}_{\text{con}} = -\frac{1}{|\{z^{+}\}|} \log \left( \frac{\sum_{z' \in \{z^{+}\}} \exp(\text{sim}(z, z') / \tau)}
{\sum_{z' \in \{z^{+}\} \cup \{z^{-}\}} \exp(\text{sim}(z, z') / \tau)} \right)
\end{equation}
where $z$ and $z'$ are feature representations extracted by $g_{\Theta}$,  
$\text{sim}(z, z') = \frac{z \cdot z'}{\|z\| \|z'\|}$ is the cosine similarity function.  
For a given input $z$, $\{z^{+}\}$ denotes the set of positive samples, while $\{z^{-}\}$ represents the set of negative samples.  
The term $|\{z^{+}\}|$ indicates the cardinality of $\{z^{+}\}$.  
$\tau$ is a temperature parameter, a positive scalar that scales the similarity scores \cite{temperature}.

The denominator of $\mathcal{L}_{\text{con}}$ plays a crucial role in distinguishing positive and negative pairs.  
If the data transformation used to generate augmented negative pairs is limited to a single type, as seen in previous studies \cite{novelty_cv2, UNODE}, the model primarily learns to differentiate positive pairs from augmented data based on that specific transformation.  
This constraint reduces the model's ability to generalize across the diverse variations commonly found in sensor-based smart environments \cite{dynamics, group_dynamics, data_shift, TFC}.  
To overcome this limitation, we design the method to incorporate multiple types of negative pairs, allowing the model to compare positive pairs with augmented data from various transformations simultaneously.  
This approach mitigates overfitting to specific patterns and enhances the model's ability to extract robust invariant representations.

\subsubsection{Invariant Representation Learning.}

To extract robust representations resilient to various types of variations in human activities using sensors, negatives generated from the \textit{Customized Negative Data Generator} based on $CST$ are utilized in contrastive learning and auxiliary classification.  
By \textit{pushing away} different types of negative samples, \texttt{CLAN} filters out insignificant variations and learns key representations that preserve essential patterns of known activities.  
Furthermore, by \textit{pulling together} samples augmented from the same data transformation method, \texttt{CLAN} forms multiple clusters in the latent space, each capturing invariant properties of known activities from different perspectives.  
This enables \texttt{CLAN} to learn invariant representations of known activities that are \textit{discriminative} against new activities.

For a given \( j \)-th data transformation method from \( CST \), the \textit{projection head}-based representations \( z_{i(j)}=g_{\Theta}(x_{i(j)}) \) and \( \Tilde{z}_{i(j)}=g_{\Theta}(\Tilde{x}_{i(j)}) \) are closer compared to samples augmented by other transformation types.  
For simplicity, we denote the identity data transformation method as \(\mathcal{T}_{0}=I \ (\text{Identity})\), which represents the original samples.  
We include this as the \textit{0}-th element of \( CST \), such that \( z_{i(0)}=z_i=g_{\Theta}(x_{i})\).  
The loss function for \texttt{CLAN}'s contrastive learning is defined as:
\begin{equation} \label{CON}
\mathcal{L}_{\textit{CON}} = - \frac{1}{B(K + 1)} \sum_{j=0}^{K} \sum_{i=1}^{B} 
\log \left( 
\frac{\exp\left(\text{sim}(z_{i(j)}, \tilde{z}_{i(j)}) / \tau\right)}{Z_{i(j)}} \right)
\end{equation}
\begin{equation*}
\begin{aligned}
Z_{i(\mathit{j})} 
& = \underbrace{\exp\left(\text{sim}(z_{i(\mathit{j})}, \tilde{z}_{i(\mathit{j})}) / \tau\right)}_{\text{Positive: Samples from the same data transformation method (\textit{j})}} \\
& + \underbrace{\sum_{k \neq j} \exp\left(\text{sim}(z_{i(\mathit{j})}, z_{i(k)}) / \tau\right) +
\exp\left(\text{sim}(z_{i(\mathit{j})}, \tilde{z}_{i(k)}) / \tau\right)}_{\text{Negative: Samples from other data transformation methods (not \textit{j})}} \\
& \quad + \underbrace{\sum_{m \neq i} \exp\left(\text{sim}(z_{i}, z_{m}) / \tau\right) + \exp\left(\text{sim}(z_{i}, \tilde{z}_{m}) / \tau\right)}_{\text{Negative: Other samples in the batch}}
\end{aligned}
\end{equation*}
where \( B \) is the batch size, and \( K \) is the cardinality of \( CST \).  
To ensure that the pairs \((\tilde{z}_{i(j)}, z_{i(j)})\) and \((z_{i(j)}, \tilde{z}_{i(j)})\) are treated equivalently as matches in Equation (\ref{contrastiveloss}), the loss terms for \(\text{sim}(\tilde{z}_{i(j)}, z_{i(j)}) / \tau\) are also incorporated into \(\mathcal{L}_{\text{CON}}\) \cite{SimCLR}.  

\texttt{CLAN} incorporates an \textit{auxiliary classifier} \cite{novelty_cv2,softmax_classifier,data_shift} to improve the separation between samples generated by different data augmentation methods.  
The goal is to classify \(K+1\) types (including the identity transformation) in \(CST\) as follows:
\begin{equation}\label{CLS}
\mathcal{L}_{CLS} =  - \frac{1}{B(K+1)}\sum_{j=0}^{K}\sum_{i=1}^{B} \left(\log(s_{i(j)}^j) + \log(\Tilde{s}_{i(j)}^j)\right)
\end{equation}
where \(s_{i(j)}^j\) and \(\Tilde{s}_{i(j)}^j\) are the predicted probabilities that \(x_{i(j)}\) and \(\Tilde{x}_{i(j)}\) belong to the \( j \)-th data transformation type in \( CST \), respectively, as computed by the softmax function.

\subsubsection{Time-Frequency Representation Learning.} \label{TF_learning}
Even with invariant representation learning, patterns may overlap between known and new activities in the temporal domain.  
Although activities differ, humans often share fundamental movements or actions across various activities.  
To address this issue, we extend the feature space by leveraging both time and frequency domains, which exhibit decomposable and generalizable characteristics across sensor datasets \cite{TFC, TF_wearable, TF_WWW3}.  
Consider two samples, \( x_1 \) and \( x_2 \), that share similar temporal patterns:
\[
x_1 = \sin(2 \pi f_1 t), \quad x_2 = \sin(2 \pi f_1 t) + \epsilon(t)
\]
where \( x_1 \) represents a waveform with frequency \( f_1 \) over time \( t \), and \( x_2 \) includes small variations \( \epsilon(t) \) relative to \( x_1 \).  
The difference between their Fourier transforms \cite{TS-analysis, fft2}, denoted as \( X_1(f) \) and \( X_2(f) \), is given by:
\begin{equation*}
|X_1(f) - X_2(f)| = \left| \int_{-\infty}^{\infty} \left(\sin(2 \pi f_1 t) - (\sin(2 \pi f_1 t) + \epsilon(t))\right) e^{-j 2 \pi f t} \, dt \right|
\end{equation*}
\begin{equation*}
= \left| \int_{-\infty}^{\infty} (-\epsilon(t)) e^{-j 2 \pi f t} \, dt \right|
\end{equation*}
where \( e^{-j 2 \pi f t} \) is the complex exponential function used in the Fourier transform to decompose time-domain data into frequency components.  
This demonstrates how the small variation \( \epsilon(t) \) in the time domain can propagate across multiple frequency components, making \( |X_1(f) - X_2(f)| \) non-negligible.  
Thus, despite similar temporal patterns between activities, they can be distinguished in the frequency domain.

Building on the insight, we design a two-tower model to learn representations for the time and frequency domains, respectively.  
The basic mechanisms for both domains are identical, except that input \( x_i \) is processed through the \textit{FFT Transformer} to be transformed into \( X_i(f) \) as the frequency domain input, which is then passed to the \textit{Frequency-wise mechanism}.  
Importantly, to extract optimized representations for each domain, we construct a separate \( CST \) for each aspect, denoted as \( CST^T \) and \( CST^F \), ensuring that both are \( \geq 2 \).  
The \textit{Customized Negative Data Generator} generates negative samples according to \( CST^T \) and \( CST^F \) and sends them to the \textit{Time-wise Encoder} \( g_{\Theta}^{T} \) and the \textit{Frequency-wise Encoder} \( g_{\Theta}^{F} \), respectively.  
We independently apply \(\mathcal{L}_{CON}\) and \(\mathcal{L}_{CLS}\) to each domain, formulated as:  
\(\mathcal{L}_{T} = \mathcal{L}_{TCON} + \mathcal{L}_{TCLS}, \mathcal{L}_{F} = \mathcal{L}_{FCON} + \mathcal{L}_{FCLS}\).  
The final objective for the overall discriminative learning module in \texttt{CLAN} is expressed as:  
\begin{equation} \label{final}  
\mathcal{L}_{\texttt{CLAN}} = \mathcal{L}_{T} + \mathcal{L}_{F}
\end{equation}

\subsection{New Activity Detection} \label{score function}

The core idea behind \texttt{CLAN}'s new activity detection is to leverage the expanded feature space generated during representation learning with \textit{CST}, allowing comparisons with known activity representations from multiple perspectives. This approach enables variation-robust new activity detection even \textit{without prior information on new activities}.  

The \textit{New Activity Detection} stage measures the similarity of \( x_{\text{test}} \) using learned representations from the \textit{Discriminative Representation Learning} stage and determines whether \( x_{\text{test}} \) corresponds to a new activity.  
First, the \textit{Similarity Calculator} utilizes representations obtained from \(\mathcal{L}_{\text{CON}}\) to compute the \textit{cosine similarity} between \( z = g_{\Theta}(x_{\text{test}}) \) and the most similar training sample, denoted as \( z_{\text{near}} \).  
Additionally, this process is applied to the negative samples of \( z \), denoted as \( z_{(j)} \), which are generated using \( CST \).  
This approach facilitates a multi-faceted analysis of the similarity between \( x_{\text{test}} \) and the training samples, expressed as:
\begin{equation} \label{sc_con}
sc_{\text{CON}} = \sum_{j=0}^{K} \text{sim}(z_{(j)}, z_{\text{near}(j)})
\end{equation}

Moreover, the \textit{Probability Calculator} employs an auxiliary classifier trained with \(\mathcal{L}_{\text{CLS}}\) to evaluate the probability \( s_{(j)}^j \) that \( z_{(j)} \) belongs to the \( j \)-th data transformation class.  
This helps determine whether \( z_{(j)} \) represents a negative sample that potentially originates from a known activity when the \( j \)-th data augmentation is applied, formalized as:
\begin{equation} \label{sc_cls}
sc_{\text{CLS}} = \sum_{j=0}^{K} s_{(j)}^{j}
\end{equation}

Finally, the new activity detection score is computed by summing both the time domain (\( sc_{T} = sc_{TCON} + sc_{TCLS} \)) and the frequency domain (\( sc_{F} = sc_{FCON} + sc_{FCLS} \)):
\begin{equation} \label{total}
sc_{\texttt{CLAN}} = sc_{T} + sc_{F}
\end{equation}
where higher values of \( sc_{\texttt{CLAN}} \) indicate a greater likelihood that \( x_{\text{test}} \) corresponds to a known activity, as elevated scores in both \( sc_{\text{CON}} \) and \( sc_{\text{CLS}} \) suggest a higher degree of alignment with extracted patterns in the training data.  

In conclusion, Algorithm~\ref{alg2} summarizes the comprehensive procedure for \texttt{CLAN}.

\begin{algorithm}[]
\caption{\texttt{CLAN} for New Activity Detection}
\label{alg2}
\RaggedRight
// \textit{Discriminative Representation Learning} \\
\textbf{Input}: Training dataset $\mathcal{X}$, 
$CST = \{\mathcal{T}_{i}\}_{i=1}^{K}$ \\
\textbf{Output}: Trained models $g_{\Theta}^{T}$ and $g_{\Theta}^{F}$ \\
\begin{algorithmic}[1]
\STATE Generate augmented dataset $\mathcal{X}_{\text{aug}}$ using all $\mathcal{T}$ in $CST$ 
\STATE Construct $\mathcal{X}_{\text{total}}=\mathcal{X} \cup \mathcal{X}_{\text{aug}}$ 
\STATE Construct $\mathcal{Y}_{\text{total}}$, 0 for $\mathcal{X}$ and $i \in \{1, \dots, K\}$ for $\mathcal{X}_{\text{aug}}$ according to the type of $\mathcal{T}$
\WHILE{not converge}
\STATE Set up batches for time domain $(\mathcal{X}_{\text{total}}, \mathcal{Y}_{\text{total}})$ and frequency domain $(FFT(\mathcal{X}_{\text{total}}), \mathcal{Y}_{\text{total}})$
\FOR{each model $g_{\Theta}^{T}$ and $g_{\Theta}^{F}$}
\STATE Extract features from the Transformer encoder
\STATE Learn invariant representation with $\mathcal{L}_{CON}$ (Eq. (\ref{CON}))
\STATE Learn auxiliary classifier with $\mathcal{L}_{CLS}$ (Eq. (\ref{CLS}))
\ENDFOR
\STATE Calculate total loss of \texttt{CLAN} with $\mathcal{L}_{\texttt{CLAN}}$ (Eq. (\ref{final}))
\STATE Update parameters of $g_{\Theta}^{T}$ and $g_{\Theta}^{F}$ using an optimizer
\ENDWHILE
\end{algorithmic}
// \textit{New Activity Detection}\\
\textbf{Input}: Trained models $g_{\Theta}^{T}$ and $g_{\Theta}^{F}$, test dataset $\mathcal{X}^{\text{test}}$, $CST$ \\
\textbf{Output}: New activity detection performance results on $\mathcal{X}^{\text{test}}$ \\
\begin{algorithmic}[1]
\STATE Construct ($\mathcal{X}_{\text{total}}^{\text{test}}$, $\mathcal{Y}_{\text{total}}^{\text{test}}$) as in training
\FOR{each $(x, y) \in (\mathcal{X}_{\text{total}}^{\text{test}}, \mathcal{Y}_{\text{total}}^{\text{test}})$}
\STATE Generate $x^{F} = FFT(x)$ for the frequency domain
\FOR{each domain ($x$, $g_{\Theta}^{T}$ for time, $x^F$, $g_{\Theta}^{F}$ for frequency)}
\STATE Compute similarity and norm values using learned invariant representations to get $sc_{CON}$ (Eq. (\ref{sc_con}))
\STATE Compute predicted probability by the auxiliary classifier to get $sc_{CLS}$ (Eq. (\ref{sc_cls}))
\ENDFOR
\STATE Calculate total new activity detection score $sc_{\texttt{CLAN}}$ (Eq. (\ref{total}))
\ENDFOR
\STATE Calculate new activity detection performance using all $sc_{\texttt{CLAN}}$
\STATE \textbf{return} New activity detection performance
\end{algorithmic}
\end{algorithm}

%% file: content/4.setup.tex
\section{Experimental Setup}
In this section, we provide the experimental setup and implementation details for evaluating the performance of \texttt{CLAN}, covering datasets, pre-processing, baselines, implementation specifics, and evaluation metrics.





\begin{figure*}[h]
    \centering
    \includegraphics[width=\textwidth]{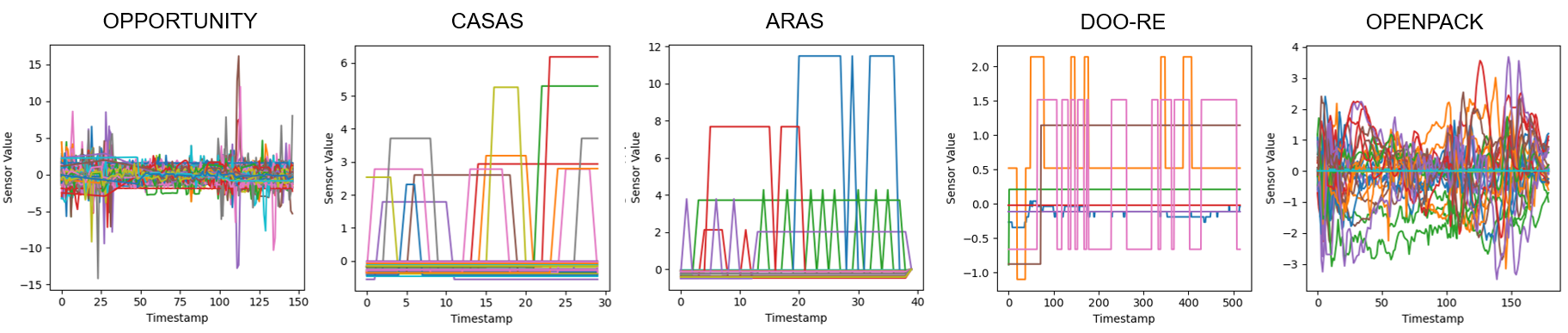}
    \caption{Visualization of examples for each dataset. 
    The x-axis represents timestamps and the y-axis represents z-score normalized sensor values.}
    \label{dataset}
\end{figure*}

\subsection{Datasets and Pre-processing}
To evaluate the generalization ability of \texttt{CLAN} in detecting new activities, we conduct extensive experiments on \textit{five} complex human activity datasets \cite{complex_activity} within sensor-based smart environments \cite{HAR_survey_3}, covering different properties of environmental domains and activity types, as well as various sensor types, as exemplified in Fig. \ref{dataset}.

\begin{itemize}
\item \textbf{\textit{OPPORTUNITY}} \cite{opportunity} is designed for the recognition of five daily activities (e.g., coffee time), with each of the four subjects performing them individually in a \textbf{controlled indoor environment}.
It includes 242 wearable and ambient sensors (e.g., 3D acceleration sensors in drawers) and is sampled at 1Hz.  The dataset contains a total of 120 activity instances.

\item \textbf{\textit{CASAS}} \cite{casas} consists of 51 \textit{motion} sensors, 2 \textit{item} sensors, and 8 \textit{door} sensors deployed in a \textbf{smart home} environment to capture 14 distinct multi-user daily activities (e.g., packing supplies in a picnic basket).  
The dataset contains a total of 469 activity instances.

\item \textbf{\textit{ARAS}} \cite{aras} captures 16 simple daily activities (e.g., watching TV) in a multi-resident \textbf{smart home}, using 20 types of ambient sensors (e.g., door sensors).  
The dataset contains a total of 3,088 activity instances.

\item \textbf{\textit{DOO-RE}} \cite{doo-re} contains data collected using seven types of ambient sensors (e.g., seat occupation sensors) in a \textbf{university seminar room}, with at least three participants performing four different group activities (e.g., seminar).  
The dataset contains 340 activity instances.

\item \textbf{\textit{OPENPACK}} \cite{openpack} involves 10 packing process activities (e.g., relocating product labels). Data is collected using four \textit{IMU} sensors, two \textit{E4} sensors, and two types of \textit{sensor device} sensors in an \textbf{industrial setting}.  
This dataset samples data every 33 Unix timestamps and includes 19,506 activity instances.
\end{itemize}

To address privacy concerns and mimic real-world scenarios, all \textit{user-identifiable information is excluded} from the datasets.
Details about the activities and sensor types within each dataset are described in Appendix \ref{A. dataset_details}.

The raw sensor data collected from these datasets undergoes pre-processing.
In this paper, we leverage widely adopted techniques from signal processing and HAR, such as signal smoothing for continuous-type data, sensor value quantization for discrete-type data, and z-score normalization for all-type data.
For activity segmentation, building on previous research \cite{changepoint, HAR_survey_1} and aiming to capture long-term correlations within entire activity sequences, we utilize \textit{change point detection method}s to segment each activity episode as a sample instance within the sensor streaming data, resulting in samples of varying lengths. 
To enable Transformer \cite{transformer}-based representation learning with these variable-length samples, we apply padding techniques commonly used in other research domains.
Through this process, the refined training dataset denoted as $\mathcal{X} \in \mathbb{R}^{N \times D \times L}$ is prepared.

\subsection{Baselines} \label{base}
\texttt{CLAN} is compared with \textit{eight} representative novelty detection baselines in unsupervised settings, \textit{without utilizing any new activity data during training}.
These baselines include \textit{Classification-based approaches:}
\begin{itemize} 
\item \textbf{\textit{OC-SVM}} \cite{ocsvm}, which employs kernel functions to find the optimal hyperplane for known class samples. 
\item \textbf{\textit{DeepSVDD}} \cite{one_class}, which learns a hypersphere boundary in the feature space that encapsulates features of known class samples. 
\end{itemize}
and \textit{Reconstruction-based methods:}
\begin{itemize} \item \textbf{\textit{AE}} \cite{OCAE, ocgan}, which utilizes an autoencoder to identify new class samples when the reconstruction error exceeds a predefined threshold. \item \textbf{\textit{OC-GAN}} \cite{ocgan}, which integrates generative adversarial networks (GANs) to generate synthetic data and detect novelty through reconstruction errors. \end{itemize}
Additionally, we incorporate recent \textit{Self-supervised novelty detection techniques}:
\begin{itemize} \item \textbf{\textit{SimCLR}} \cite{SimCLR, novelty_cv2}, which trains an encoder to learn representations of known classes by maximizing the similarity between original and augmented samples, utilizing a similarity score for novelty detection. \item \textbf{\textit{GOAD}} \cite{GOAD}, which optimizes a feature space through data augmentations to ensure inter-class distances remain larger than intra-class distances, with novelty likelihood determined by the distance from cluster centers. \item \textbf{\textit{FewSOME}} \cite{FewSome}, which employs Siamese networks with shared weights to construct closely proximate representations and incorporates the \textit{Stop Loss} mechanism to prevent representational collapse, detecting novelty based on the distance of learned representations. 
\item \textbf{\textit{UNODE}} \cite{UNODE}, which adopts a probabilistic contrastive learning approach by leveraging the Kullback-Leibler divergence to generate negative pairs, utilizing a similarity score mechanism to distinguish new class instances. \end{itemize}

For both \texttt{CLAN} and the self-supervised baselines, 10 commonly used data augmentation methods for HAR and time-series sensor data \cite{tsaug, TFC, ts-tcc, TF_wearable} are leveraged, as shown in Table \ref{tab: data augmentation methods}, including \textit{AddNoise}, \textit{Convolve}, \textit{Permute}, \textit{Drift}, \textit{Dropout}, \textit{Pool}, \textit{Quantize}, \textit{Scale}, \textit{Reverse}, and \textit{TimeWarp}.

\begin{table} [h]
\centering
\caption{Description and implementation details of data augmentation methods used in this paper.
Words enclosed within \textit{italics} signify parameters that can be employed with each data augmentation method.
Values enclosed in parentheses denote specific experimental values used during the experiments and they are drawn from commonly utilized configurations in previous studies or libraries.
}
\scalebox{0.9}{
\begin{tabular}{c|p{5.2in}}
\hline
\label{tab: data augmentation methods}
\textbf{Types}&\textbf{Description and implementation details}\\\hline
AddNoise
& Injects random Gaussian noise into the input time series data, with the noise's intensity scaled by a factor of \textit{scale\_num} (0.01).\\\hline
Convolve
& Performs a convolution operation on the input time series data using a specified \textit{kernel window type} (flattop) and a \textit{window\_size} (11).
\\\hline
Permute
& Segments the time series data into a variable number of sections between \textit{min\_segments} (1) and \textit{max\_segments} (5) and then reshuffles these segments.
\\\hline
Drift
& 
Gradually shifts the values of the time series data, with the maximum shift magnitude controlled by \textit{max\_drift} (0.7), and the number of affected data points determined by \textit{n\_drift\_points} (5).\\\hline
Dropout
& Randomly omits data points in the time series with a probability of \text{p} (10\%) and replaces them with a specified \text{fill} value (0). \\\hline
Pool
& Aggregates time series data within non-overlapping windows of \textit{size} (4) through a pooling operation, thereby downsampling the time series by selecting representative values from each window. \\\hline
Quantize
& Discretizes individual data points within the time series into one of a defined number of discrete \textit{n\_levels} (20).
\\\hline
Scale
& Adjusts each data point's value by applying random scaling factors drawn from a normal distribution centered around \textit{loc} (2) with a standard deviation of \textit{sigma} (1.1).
\\\hline
Reverse
& Inverts the temporal order of data points, effectively flipping the sequence of the time series data.\\\hline
TimeWarp
& Modifies the time series data's tempo by introducing up to \textit{n\_speed\_change} (5) alterations, where the time axis can be stretched or compressed up to a specified \textit{max\_speed\_ratio} (3).
\\ 
\hline
\end{tabular}
}
\vspace{-10pt}
\end{table}

Each approach employs distinct mechanisms for representation learning and new pattern detection, and the experiments are conducted according to the specific mechanisms of each approach. 
For the baselines, either the backbone architecture from the original papers or the Transformer encoder \cite{transformer} used in this work is selected, based on which yields better results. 
To ensure a fair and consistent comparison, the same training strategy is applied across all experiments for both the baselines and \texttt{CLAN}. 

\subsection{Implementation Details}
Experiments are conducted on a server equipped with an NVIDIA TITAN RTX GPU, an Intel Xeon Gold 5215 CPU (2.5GHz), 256GB RAM, and Ubuntu 18.04.5 LTS. 
The implementation of \texttt{CLAN} is based on Anaconda 4.10.1, Python 3.7.16, and PyTorch 1.12.1. The frequency domain representation of each sample is computed using PyTorch’s FFT library.  
The code and experimental details are publicly accessible on the project's repository \footnote{\url{https://github.com/cdsnlab/CLAN}}.

Recent advancements in HAR or time series sensor data domains \cite{HAR_survey_4, ART, IF-ConvTrans, THAT} have showcased the efficacy of Transformer \cite{transformer}-variant methods inspired by NLP tasks. 
We employ the Transformer encoder \cite{transformer} as $g_{\Theta}$, which is employed as the backbone and connects the projection layer and linear layer on top. 
The backbone consists of a stack of two encoder layers ($M$ =2), where each layer has one attention head and a feedforward neural network with a dimension that is twice the length of the input data. 
The projection layer is responsible for extracting representations $z_{i}$ to facilitate contrastive learning ($\mathcal{L}_{CON}$), and the linear layer is tasked with extracting representations $s_{i}$ for classification tasks ($\mathcal{L}_{CLS}$).
\textit{The projection layer} is composed of two fully connected (linear) layers along with batch normalization and a ReLU activation function, to convert the extracted backbone features into 128-dimensional embedding feature vectors \cite{TFC}.
\textit{The linear layer} is a single layer that scales the feature output in $(K+1)$ dimensions to match the number of data transformation method types.

To evaluate performance robustness, experiments are performed with 10 different random seeds, and the average performance is presented. 
The dataset is split into training, validation, and test sets in a 60:20:20 ratio, where the training set contains only known activity samples, while both known and new activities are included in the validation and test sets. 
Training is conducted using a mini-batch approach with a batch size of \( B=64 \) over 100 epochs. 
The encoders are optimized using Adam with a learning rate of \( 3 \times 10^{-4} \), and the loss function is configured with \( \tau = 0.5 \). 
The values of \( B \), learning rate, and \( \tau \) are determined based on the average best performance across the datasets.  
The threshold \( \theta_{CST} \) is selected from a predefined AUROC threshold set \{0.5, 0.6, 0.7, 0.8, 0.9\}. 
Depending on the known activity types within each dataset, \( \theta_{CST} \) is set to \textit{the highest value from which at least two distinct data transformation methods can be derived}.

\subsection{Evaluation Metrics}
To comprehensively evaluate \texttt{CLAN}'s effectiveness, we employ widely used metrics for assessing novelty detection performance: \textbf{AUROC} and \textbf{Balanced Accuracy (ACC)}.
Higher values indicate better performance.
\begin{itemize}
    \item[$-$] \textit{\textbf{AUROC}} serves as a comprehensive indicator of the model's capability to differentiate between known and new activity samples across varying detection thresholds.
    \item[$-$] \textit{\textbf{Balanced Accuracy}}, defined as $\frac{\text{sensitivity} + \text{specificity}}{2}$ \cite{FewSome}, provides an unbiased evaluation of detection performance for both known and new activity classes.  
    The predicted label for each sample is assigned based on the percentile of its new activity detection score.
\end{itemize}

%% file: content/5.results.tex
\section{Evaluation Results}
In this section, we evaluate the new activity detection performance of \texttt{CLAN} through extensive quantitative and qualitative experiments.
\subsection{New Activity Detection Performance} \label{q1}

\begin{table}[b]
\centering
\caption{
New activity detection performance (\%) ($\pm$standard deviation) for five human activity datasets in \textit{one-class} and \textit{unsupervised} setups. 
For each metric, the best-performing value is highlighted in bold and the second-best value is underlined.
}
\label{tab 2: one-class}
\scalebox{0.75}{
\begin{tabular}{c|cc|cc|cc|cc|cc}
\hline
\multicolumn{1}{c}{} 
& \multicolumn{2}{c}{\textbf{OPPORTUNITY}}
& \multicolumn{2}{c}{\textbf{CASAS}} 
& \multicolumn{2}{c}{\textbf{ARAS}}
& \multicolumn{2}{c}{\textbf{DOO-RE}} 
& \multicolumn{2}{c}{\textbf{OPENPACK}}
\\
\hline
\textbf{Model} & \textbf{AUROC $\uparrow$}  & \textbf{ACC $\uparrow$}  
& \textbf{AUROC $\uparrow$}  & \textbf{ACC $\uparrow$} 
& \textbf{AUROC $\uparrow$}  & \textbf{ACC $\uparrow$}    
& \textbf{AUROC $\uparrow$}  & \textbf{ACC $\uparrow$} 
& \textbf{AUROC $\uparrow$}  & \textbf{ACC $\uparrow$} 
\\
\hline
{OC-SVM} 
& \Large82.5 & \Large71.8 
& \Large80.8 & \Large67.9 
& \Large72.0 & \Large59.2 
& \Large{50.2} & \Large50.1 
& \Large62.3 & \Large57.1 
\\
\cite{ocsvm}
& $\pm$2.0 & $\pm$1.5 
& $\pm$14.7 & $\pm$13.0 
& $\pm$7.8 & $\pm$6.5 
& $\pm$2.3 & $\pm$1.5
& $\pm$11.4 & $\pm$7.0 
\\
{DeepSVDD}
& \Large86.1 & \Large81.9
& \Large63.4 & \Large60.3 
& \Large68.3 & \Large59.7
& \Large62.6 & \Large60.6 
& \Large57.5 & \Large53.1 
\\
\cite{one_class}
& $\pm$2.4 & $\pm$2.2 
& $\pm$18.8 & $\pm$9.3 
& $\pm$12.7 & $\pm$4.8
& $\pm$4.4 & $\pm$1.2 
& $\pm$8.2 & $\pm$4.0 
\\\hline
{AE}
& \Large62.2 & \Large55.2 
& \Large55.7 & \Large51.6 
& \Large57.9 & \Large51.0 
& \Large76.8 & \Large66.7 
& \Large57.6 & \Large52.4 
\\
\cite{OCAE}
& $\pm$21.7 & $\pm$20.2 
& $\pm$18.8 & $\pm$6.6 
& $\pm$11.9 & $\pm$2.6 
& $\pm$10.3 & $\pm$12.0 
& $\pm$20.9 & $\pm$7.9 
\\
{OCGAN}
& \Large61.5 & \Large53.3 
& \Large60.6 & \Large53.8 
& \Large60.1 & \Large52.0 
& \Large77.9 & \Large67.8 
& \Large55.7 & \Large52.5
\\
\cite{ocgan}
& $\pm$20.7 & $\pm$20.5 
& $\pm$19.3 & $\pm$8.0 
& $\pm$12.2 & $\pm$3.5 
& $\pm$6.2 & $\pm$6.3 
& $\pm$17.0 & $\pm$5.9 
\\\hline
SimCLR
& \underline{\Large90.7} & \underline{\Large87.7} 
& \Large58.3 & \Large60.0 
& \Large77.7 & \Large69.6 
& \Large63.6 & \Large61.7 
& \Large76.6 & \Large67.3 
\\
\cite{SimCLR}
& $\pm$2.7 & $\pm$5.6 
& $\pm$13.0 & $\pm$6.6 
& $\pm$7.3 & $\pm$5.4 
& $\pm$12.3 & $\pm$8.5 
& $\pm$8.0 & $\pm$6.1 
\\
GOAD
& \Large63.0 & \Large60.1
& \Large78.9 & \Large66.0 
& \underline{\Large84.8} & \Large66.7 
& \Large71.5 & \Large64.5 
& \underline{\Large82.3} & \underline{\Large67.7} 
\\
\cite{GOAD}
& $\pm$4.9 &$\pm$4.2 
& $\pm$7.1 & $\pm$4.0 
& $\pm$5.3 & $\pm$6.4
& $\pm$5.8 & $\pm$5.4 
& $\pm$6.5 & $\pm$5.3 
\\
FewSome
& \Large88.7 & \Large84.7 
& \underline{\Large91.1} & \underline{\Large79.5} 
& \Large81.6 & \Large68.9 
& \underline{\Large91.4} & \underline{\Large83.0} 
& \Large61.3 & \Large56.1 
\\
\cite{FewSome}
& $\pm$4.3 &$\pm$5.6 
&$\pm$6.3 & $\pm$7.3 
& $\pm$9.2 & $\pm$6.5
& $\pm$4.7 & $\pm$4.4 
& $\pm$6.6 & $\pm$3.8 
\\
UNODE
& \Large85.0 & \Large84.1 
& \Large65.0 & \Large74.3 
& \underline{\Large84.8} & \underline{\Large72.6} 
& \Large78.8 & \Large70.0 
& \Large73.3 & \Large64.1 
\\
\cite{UNODE}
& $\pm$12.0 &$\pm$8.8 
&$\pm$31.2 & $\pm$16.5 
& $\pm$17.5 & $\pm$12.4
& $\pm$3.8 & $\pm$3.6 
& $\pm$10.2 & $\pm$6.4 
\\
\hline
\hline
\textbf{\texttt{CLAN}}
& \textbf{\Large98.0} & \textbf{\Large95.6} 
& \textbf{\Large98.7} & \textbf{\Large93.2} 
& \textbf{\Large94.9} & \textbf{\Large82.4} 
& \textbf{\Large95.8} & \textbf{\Large91.0} 
& \textbf{\Large93.0} & \textbf{\Large83.4} 
\\
& $\pm$2.0 &$\pm$3.8 
&$\pm$1.5 & $\pm$5.0 
& $\pm$4.8 & $\pm$7.8
& $\pm$3.2 & $\pm$5.2 
& $\pm$5.3 & $\pm$7.7 
\\
\hline
\end{tabular}
}
\end{table}

\subsubsection{One-class Setup.}
Following the settings of the existing novelty detection baselines, one activity is designated as a known activity, and the other activities are regarded as new activities.
If there are $C$ activities in the dataset, $C$ distinct one-class classification tasks are performed.
Table~\ref{tab 2: one-class} presents the overall results in one-class classification scenarios, where \texttt{CLAN} outperforms other methods in discovering new patterns across different sensor-based smart environments.  

\texttt{CLAN} improves AUROC by margins of 8.0\% for \textit{OPPORTUNITY}, 8.4\% for \textit{CASAS}, 11.9\% for \textit{ARAS}, 4.8\% for \textit{DOO-RE}, and 13.0\% for \textit{OPENPACK}, respectively, surpassing the best results from the baselines.  
While existing baselines exhibit considerable performance variability across datasets, \texttt{CLAN} consistently achieves over 90\% AUROC in all cases.  
This indicates that \texttt{CLAN} effectively extracts invariant representations that generalize well across diverse sensor-based environments.  

On average, \texttt{CLAN} enhances balanced accuracy by 9.1\% for \textit{OPPORTUNITY}, 17.2\% for \textit{CASAS}, 13.5\% for \textit{ARAS}, 9.7\% for \textit{DOO-RE}, and 23.3\% for \textit{OPENPACK}, respectively, surpassing the best baseline performances.  
\texttt{CLAN} achieves superior balanced performance in identifying both known and new activities compared to other baselines, effectively minimizing feature overlap between known and new activities and demonstrating its ability to derive discriminative representations.  

The standard deviation across \( C \) distinct one-class classification tasks is relatively smaller for \texttt{CLAN} compared to the baselines, indicating that \texttt{CLAN} performs stably regardless of activity type.  
This demonstrates that \texttt{CLAN}'s customized and multi-perspective representation-based mechanism is effective across different datasets as well as for the unique characteristics of each activity.  

Self-supervised types of methods outperform other approaches, demonstrating that contrast mechanisms for instance discrimination \cite{ILD} effectively identify novel patterns despite diverse temporal dynamics in human activity.  
Among them, \texttt{CLAN} refines the feature space, improving new activity detection, whereas other baselines primarily focus on attracting positive samples while considering only a limited type of negatives, leading to less distinct representations of known activities.

\begin{table}[b]
\centering
\caption{
New activity detection performance (\%) ($\pm$standard deviation) for five human activity datasets in \textit{multi-class} and \textit{unsupervised} setups. 
For each metric, the best-performing value is highlighted in bold and the second-best value is underlined.
}
\label{multi-class}
\scalebox{0.75}{
\begin{tabular}{c|cc|cc|cc|cc|cc}
\hline
\multicolumn{1}{c}{} 
& \multicolumn{2}{c}{\textbf{OPPORTUNITY}}
& \multicolumn{2}{c}{\textbf{CASAS}} 
& \multicolumn{2}{c}{\textbf{ARAS}}
& \multicolumn{2}{c}{\textbf{DOO-RE}} 
& \multicolumn{2}{c}{\textbf{OPENPACK}}
\\
\hline
\textbf{Model} & \textbf{AUROC $\uparrow$}  & \textbf{ ACC $\uparrow$}  
& \textbf{AUROC $\uparrow$}  & \textbf{ ACC $\uparrow$} 
& \textbf{AUROC $\uparrow$}  & \textbf{ ACC $\uparrow$} 
& \textbf{AUROC $\uparrow$}  & \textbf{ ACC $\uparrow$} 
& \textbf{AUROC $\uparrow$}  & \textbf{ ACC $\uparrow$} 
\\
\hline
OC-SVM
& \Large74.3 & \Large65.9 
& \Large58.4 & \Large56.3 
& \Large52.0 & \Large51.7 
& \Large51.7 & \Large51.4 
& \Large54.2 & \Large53.1 
\\
& $\pm$4.5 & $\pm$3.3 
& $\pm$9.4 & $\pm$6.2 
& $\pm$6.8 & $\pm$4.7 
& $\pm$3.4 & $\pm$2.2 
& $\pm$6.4 & $\pm$4.7 
\\
DeepSVDD
& \Large88.6 & \Large79.3 
& \Large69.8 & \Large64.3 
& \Large53.2 & \Large52.2 
& \Large64.1 & \Large61.0 
& \Large50.0 & \Large50.0 
\\
& $\pm$3.6 & $\pm$5.0 
& $\pm$5.5 & $\pm$4.1 
& $\pm$1.4 & $\pm$1.0 
& $\pm$13.8 & $\pm$11.2 
& $\pm$0.7 & $\pm$0.6 
\\
\hline
AE
& \Large52.2 & \Large54.7 
& \Large54.4 & \Large53.4 
& \Large51.3 & \Large51.1 
& \Large67.2 & \Large64.6 
& \Large57.2 & \Large54.9 
\\
& $\pm$15.4 & $\pm$11.2 
& $\pm$9.4 & $\pm$7.3 
& $\pm$6.0 & $\pm$4.5 
& $\pm$21.2 & $\pm$18.0 
& $\pm$14.1 & $\pm$10.8 
\\
OCGAN
& \Large54.6 & \Large57.8 
& \Large54.0 & \Large53.7 
& \Large51.3 & \Large51.1 
& \Large73.6 & \Large68.3 
& \Large55.4 & \Large54.0 
\\
& $\pm$12.7 & $\pm$7.9 
& $\pm$9.2 & $\pm$7.0 
& $\pm$5.5 & $\pm$4.3 
& $\pm$16.7 & $\pm$14.7
& $\pm$10.8 & $\pm$8.0 
\\
\hline
SimCLR
& \Large92.9 & \Large85.9 
& \Large82.3 & \Large77.1 
& \underline{\Large74.9} & \underline{\Large68.5} 
& \Large73.8 & \Large67.7 
& \Large66.7 & \Large62.2 
\\
& $\pm$6.3 & $\pm$8.1 
& $\pm$18.2 & $\pm$16.8 
& $\pm$11.1 & $\pm$9.3
& $\pm$18.8 & $\pm$16.4 
& $\pm$10.4 & $\pm$8.2 
\\
GOAD
& \Large53.5 & \Large52.6 
& \Large82.8 & \Large75.8 
& \Large68.1 & \Large63.0 
& \Large71.4 & \Large63.5 
& \Large60.6 & \Large57.6 
\\
& $\pm$10.6 & $\pm$7.1 
& $\pm$5.6 & $\pm$4.9 
& $\pm$6.6 & $\pm$5.1 
& $\pm$13.1 & $\pm$13.0 
& $\pm$8.2 & $\pm$5.8 
\\
FewSome
& \Large68.4 & \Large60.9 
& \Large66.2 & \Large61.4 
& \Large56.4 & \Large54.4 
& \underline{\Large84.2} & \underline{\Large77.1} 
& \Large52.6 & \Large51.6 
\\
& $\pm$8.2 & $\pm$9.4 
& $\pm$6.3 & $\pm$6.2 
& $\pm$3.9 & $\pm$2.3 
& $\pm$6.8 & $\pm$7.5 
& $\pm$2.9 & $\pm$2.0 
\\
UNODE
& \underline{\Large93.6} & \underline{\Large87.5} 
& \underline{\Large90.2} & \underline{\Large81.9} 
& \Large72.4 & \Large66.4 
& \Large81.8 & \Large76.2 
& \underline{\Large71.6} & \underline{\Large65.6} 
\\
& $\pm$3.8 & $\pm$5.8 
& $\pm$6.7 & $\pm$7.3 
& $\pm$6.0 & $\pm$4.8 
& $\pm$11.9 & $\pm$12.1 
& $\pm$9.3 & $\pm$7.7 
\\
\hline
\hline
\texttt{\textbf{CLAN}}
& \textbf{\Large98.8} & \textbf{\Large94.3}
& \textbf{\Large95.7} & \textbf{\Large88.4} 
& \textbf{\Large77.9} & \textbf{\Large71.5}
& \textbf{\Large96.3} & \textbf{\Large90.0} 
& \textbf{\Large79.9} & \textbf{\Large72.0} 
\\
& $\pm$1.8 & $\pm$5.4 
& $\pm$1.9 & $\pm$3.2 
& $\pm$4.9 & $\pm$4.0 
& $\pm$2.8 & $\pm$4.0 
& $\pm$5.5 & $\pm$4.6 
\\
\hline
\end{tabular}
}
\end{table}

\subsubsection{Multi-class Setup.}
Inspired by previous work \cite{novelty_cv2}, we experiment with multiple types of activities present in both known and new activity sets.  
This setting better reflects real-world scenarios.  
Activities within the same dataset are split into two halves: one representing known activities and the other representing new activities.  
We conduct 10 trials using randomly selected sets of known and new activities and report the average results.  

As shown in Table~\ref{multi-class}, \texttt{CLAN} outperforms the highest baseline results, achieving greater improvements in average AUROC and balanced accuracy by margins of 5.5\% and 7.7\% for \textit{OPPORTUNITY}; 6.1\% and 7.8\% for \textit{CASAS}; 3.9\% and 4.4\% for \textit{ARAS}; 14.3\% and 16.7\% for \textit{DOO-RE}; and 11.6\% and 9.7\% for \textit{OPENPACK}.  
Despite the challenging multi-class setup, where known and new activity patterns significantly overlap, \texttt{CLAN} effectively extracts discriminative representations, maintaining superior performance.  

Compared to the high standard deviation observed in baselines, which are sensitive to the composition of known activities, \texttt{CLAN} demonstrates more stable performance across diverse known and new activity configurations.  
This stability is attributed to customized data augmentation, highlighting \texttt{CLAN}'s ability to discover invariant representations applicable across diverse configurations.

\begin{figure}[t]
    \centering
    \hfill    
    \includegraphics[width = \linewidth]{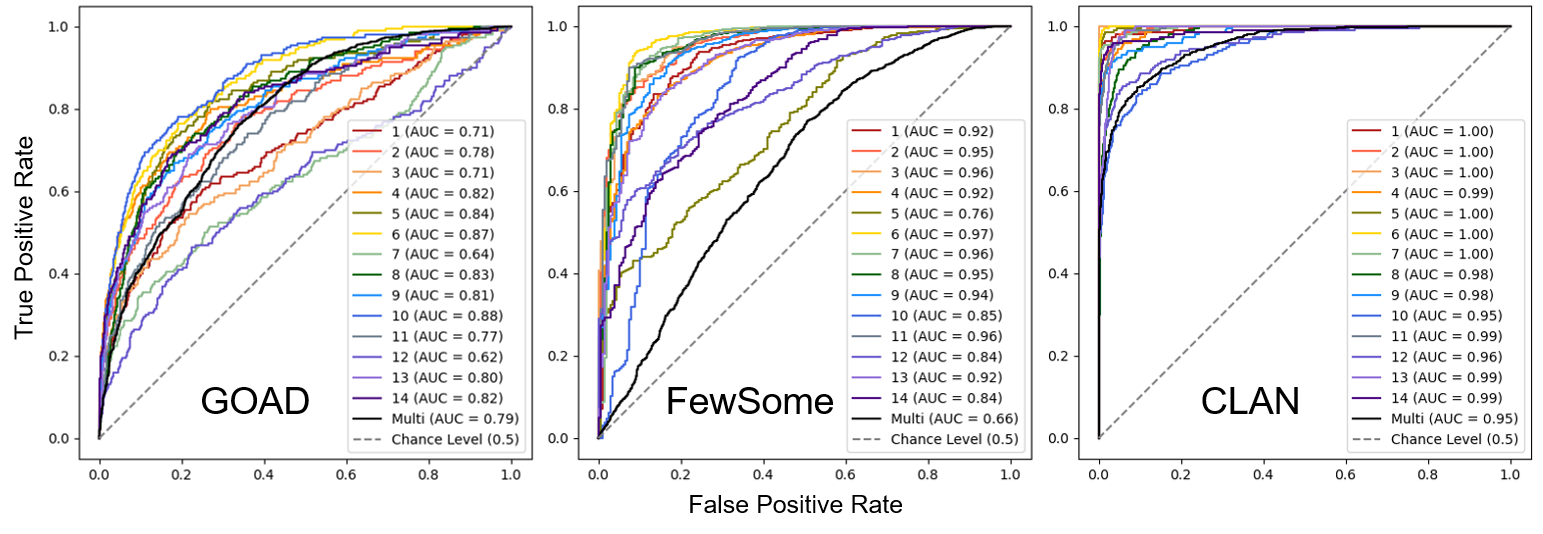}
    \caption{ROC curve graphs for \texttt{CLAN} and the top-2 best-performing baselines in \textit{CASAS}.
    ROC curves closer to the top-left corner indicate superior performance.  
    The numerical labels on each ROC curve correspond to the known activity number.  
    The \textit{Multi} label represents the performance in multi-class scenarios.
    }     
    \label{fig 7: AUROC graphs collection}
\end{figure}

\begin{figure}[t]
    \centering
    \includegraphics[width = 0.9\textwidth]{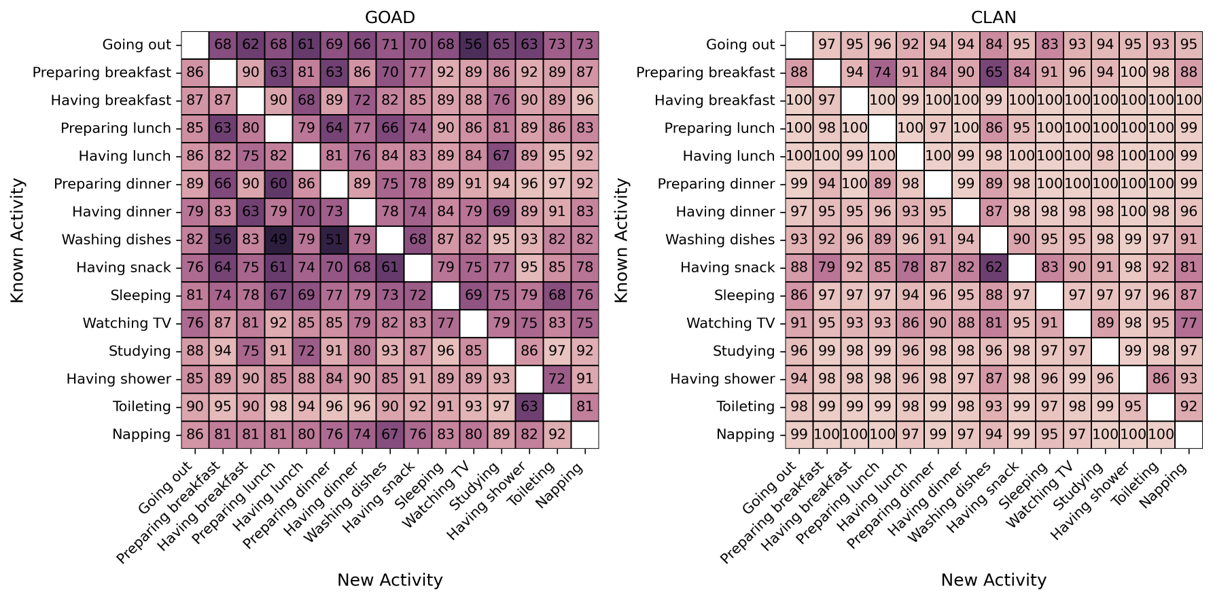}
    \caption{AUROC (\%) in \textit{ARAS} when one activity is designated as a known activity and another as a new activity. Brighter colors indicate better performance.}
    \label{matrix}
\end{figure}

\subsubsection{Class-wise Analysis.} 
To further evaluate \texttt{CLAN}'s performance on each known activity in detail, we provide a fine-grained analysis using ROC curves.  
Fig.~\ref{fig 7: AUROC graphs collection} presents an example of ROC curves in \textit{CASAS}, illustrating \texttt{CLAN}'s class-wise performance compared to the top two best-performing baselines.  

Consistent with previous experiments, \texttt{CLAN} maintains superior performance across all one-class classification tasks, not only on average but also with less variance across different activities.  
Other baseline approaches exhibit significant performance degradation depending on the activity type, such as \textit{GOAD} with \textit{Retrieve dishes from a kitchen cabinet} (12) and \textit{FewSome} with \textit{Sweep the kitchen floor} (5).  
Moreover, when a known activity like \textit{Retrieve dishes from a kitchen cabinet} (12), which may share movement characteristics with other activities, is present, \texttt{CLAN} distinguishes it more effectively compared to the baselines.  
These results indicate that by setting appropriate \( CST \)s for each known activity, \texttt{CLAN} minimizes pattern overlap among activities, establishing clear representation boundaries in the latent space.  
Additionally, \texttt{CLAN} prevents meaningless temporal dynamics from being incorporated into the representations, ensuring more robust feature extraction for each type of known activity.

In addition, to investigate \texttt{CLAN}’s ability to differentiate between similar activities when designated as known and new, we conduct pairwise new activity detection experiments in \textit{ARAS}.  
In these experiments, one activity is designated as known and another as new, as shown in Fig.~\ref{matrix}.  

The overall color intensity of \texttt{CLAN} is brighter than that of the best-performing baseline, \textit{GOAD}, visually confirming that \texttt{CLAN} distinguishes similar activities more effectively than the baseline.  
For example, in the case of \textit{Having Breakfast} (Known) and \textit{Having Lunch} (New), the visual differences between the activities are shown in Fig.~\ref{fig 9: attentionmaps}, with AUROC performance of 67.6\% for the baseline and 99.2\% for \texttt{CLAN}.  
These results highlight \texttt{CLAN}’s ability to precisely capture key representations within known activities at a fine-grained level.

\subsection{Ablation Study} 
\subsubsection{Effects of the Applying Multiple Customized Data Augmentation Strategy.} \label{q3}

\begin{figure}[b]
    \centering
    \includegraphics[width=0.85\linewidth]{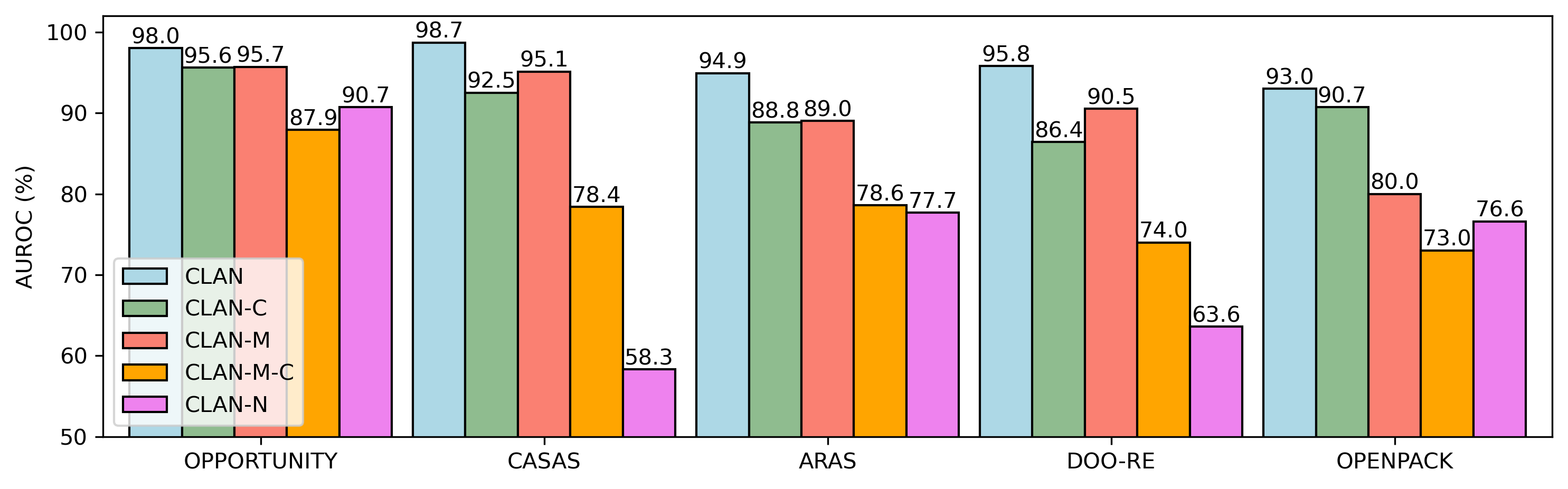}
    \caption{AUROC (\%) results for the ablation study evaluating the effectiveness of \texttt{CLAN}'s customized and multi-type data augmentation strategy.}
    \label{Effects of Diversity}
\end{figure}

Fig.~\ref{Effects of Diversity} illustrates the impact of \texttt{CLAN}'s data augmentation strategy, which incorporates customized and multiple types of strongly shifted samples as negatives.  
To evaluate its effectiveness, we compare different data augmentation strategies in representation learning and new activity detection:  
\begin{itemize}
    \item[$-$] \textit{\textbf{CLAN-C}} is non-customized, applying multiple random augmentation methods without dataset-specific adaptation.  
    \item[$-$] \textit{\textbf{CLAN-M}} is non-multiple, using only a single type of strong data transformation method derived from \( CST \).  
    \item[$-$] \textit{\textbf{CLAN-M-C}} neither applies multiple augmentations nor utilizes \( CST \); instead, it applies a single random augmentation method.  
    \item[$-$] \textit{\textbf{CLAN-N}} omits explicit negative sample generation entirely.  
\end{itemize}

(1) \texttt{CLAN} surpasses \textit{\textbf{CLAN-C}} by 6.7\%, highlighting the importance of \( CST \), and outperforms \textit{\textbf{CLAN-M}} by 4.7\%, demonstrating the critical role of customized and diverse data augmentation strategies in generating negative samples for invariant representation learning, ensuring robustness to pattern variations in activities.  
(2) Comparing \textit{\textbf{CLAN-M-C}} and \textit{\textbf{CLAN-N}} with \textit{\textbf{CLAN-C}}, \textit{\textbf{CLAN-M}}, and \texttt{CLAN}, we validate that both \textit{quantitatively} increasing the number of data augmentation methods and \textit{qualitatively} improving negatives through \( CST \) are crucial.  
(3) In a noise-sensitive environment with highly similar activities, such as \textit{OPENPACK}, the difference between \texttt{CLAN} and \textit{\textbf{CLAN-M}} suggests that multi-perspective new activity detection is particularly effective.


\begin{figure}[t] 
\centering
\subfloat[AUROC(\%) performance degradation.]{
\includegraphics[width=0.4\linewidth]{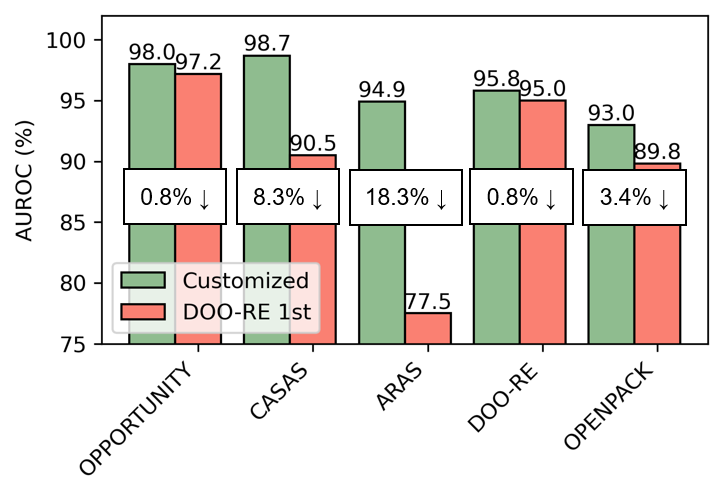}
\label{figa}
    }
\hspace{2mm}
\subfloat[t-SNE visualization.]{
\includegraphics[width=0.45\linewidth]{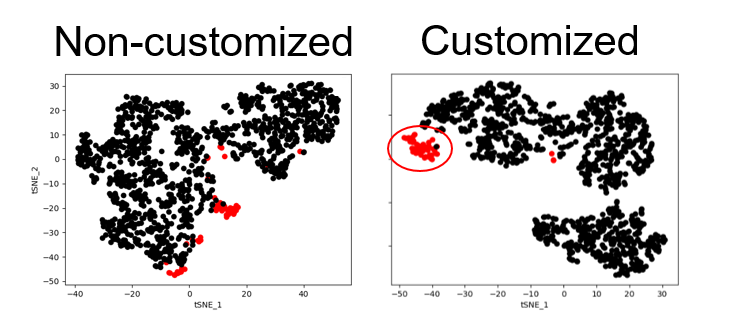}
    \label{figb}
}
\caption{Comparison of the effects of applying \texttt{CLAN}'s \( CST \) versus using the same data augmentation methods across all datasets.
(For (b), Red = known activity, Black = new activity)
}
\label{fig:customized}
\end{figure}


To investigate the effectiveness of \textit{Customized Negative Data Generation} at a fine-grained level, we conduct a case study, as depicted in Fig.~\ref{fig:customized}.  
In Fig.~\ref{figa}, \textit{Customized} refers to tailoring \( CST \) for each activity in each dataset, whereas the alternative setting applies the same transformation (\( CST \) from the first activity of \textit{DOO-RE}) to all datasets.  
This results in an overall performance drop, whether the same data transformation is applied across different datasets or to various activity types within the same dataset (\textit{DOO-RE}).  
Ignoring differences between datasets or activity types causes weakly shifted samples to be treated as negatives, leading to ambiguous boundaries of known activities in the latent space, as shown in Fig.~\ref{figb}, where \textit{Preparing lunch} in \textit{ARAS} is set as a known activity.  
These results validate the importance of customized data generation in \texttt{CLAN}.

\begin{figure}[]
    \centering
    \includegraphics[width=0.85\linewidth]{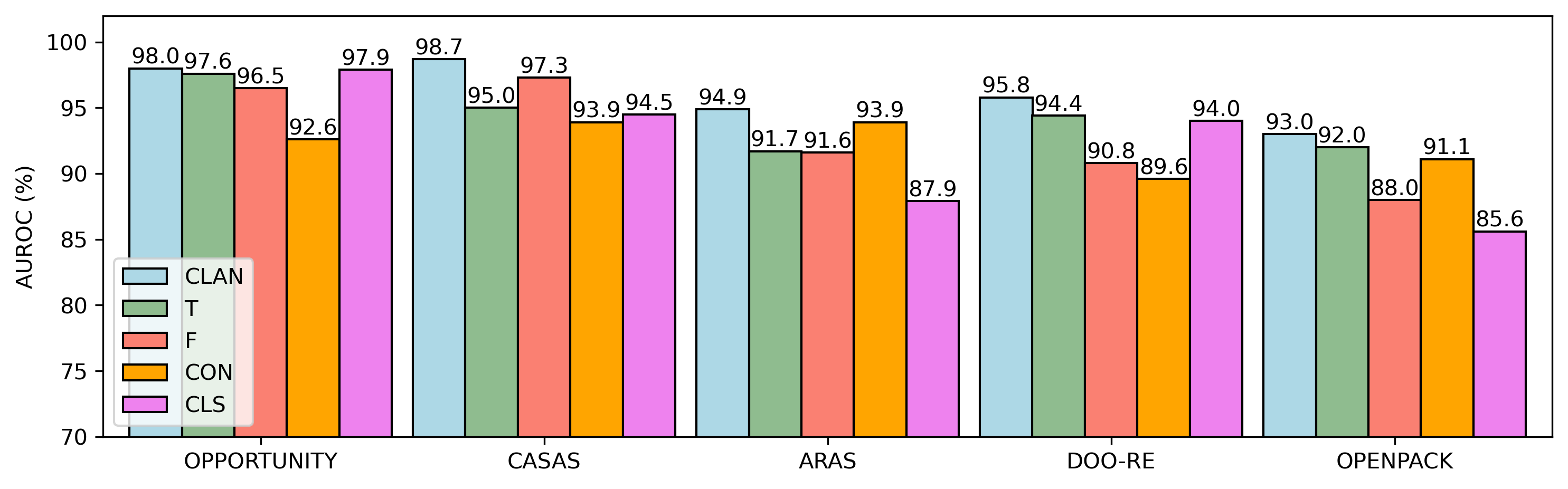}
    \caption{AUROC (\%) results from the ablation study evaluating  the effectiveness of \texttt{CLAN}'s components.}
    \label{Ablation Study}
\end{figure}

\subsubsection{Effects of Each Component.} \label{q2}

We perform an ablation study to assess the impact of different components in \texttt{CLAN} across all datasets, as shown in Fig.~\ref{Ablation Study}.  
\textbf{\textit{T} (TCON+TCLS)}, \textbf{\textit{F} (FCON+FCLS)}, \textbf{\textit{CON} (TCON+FCON)}, and \textbf{\textit{CLS} (TCLS+FCLS)} represent the results of focusing exclusively on the time domain, frequency domain, $\mathcal{L}_{CON}$, and $\mathcal{L}_{CLS}$, respectively.

(1) The results demonstrate that \texttt{CLAN}'s new activity detection performance surpasses \textbf{\textit{T}}, \textbf{\textit{F}}, \textbf{\textit{CON}}, and \textbf{\textit{CLS}} by 2.0\%, 3.5\%, 4.2\%, and 4.4\%, respectively.  
While the effectiveness of each component varies by dataset, integrating all components to extract multi-faceted discriminative representations proves effective across all datasets.  
(2) While considering temporal relations is crucial (\textbf{\textit{T}}), incorporating frequency information (\textbf{\textit{F}}) expands the feature space and enhances discrimination performance overall.  
In particular, for \textit{CASAS}, where motion sensor patterns between activities often overlap temporally, analyzing frequency helps better differentiate these patterns.  
(3) \textbf{\textit{CON}} proves highly effective in datasets with significant variation within the same known activity, such as \textit{ARAS} and \textit{OPENPACK}, as it extracts invariant representations of known activities to guard against meaningless variations.  
\textbf{\textit{CLS}} provides support in the other datasets where activities have relatively distinct features, helping to refine the discrimination boundaries of known activities.

\begin{figure}[t]
    \centering
    \includegraphics[width=\linewidth]{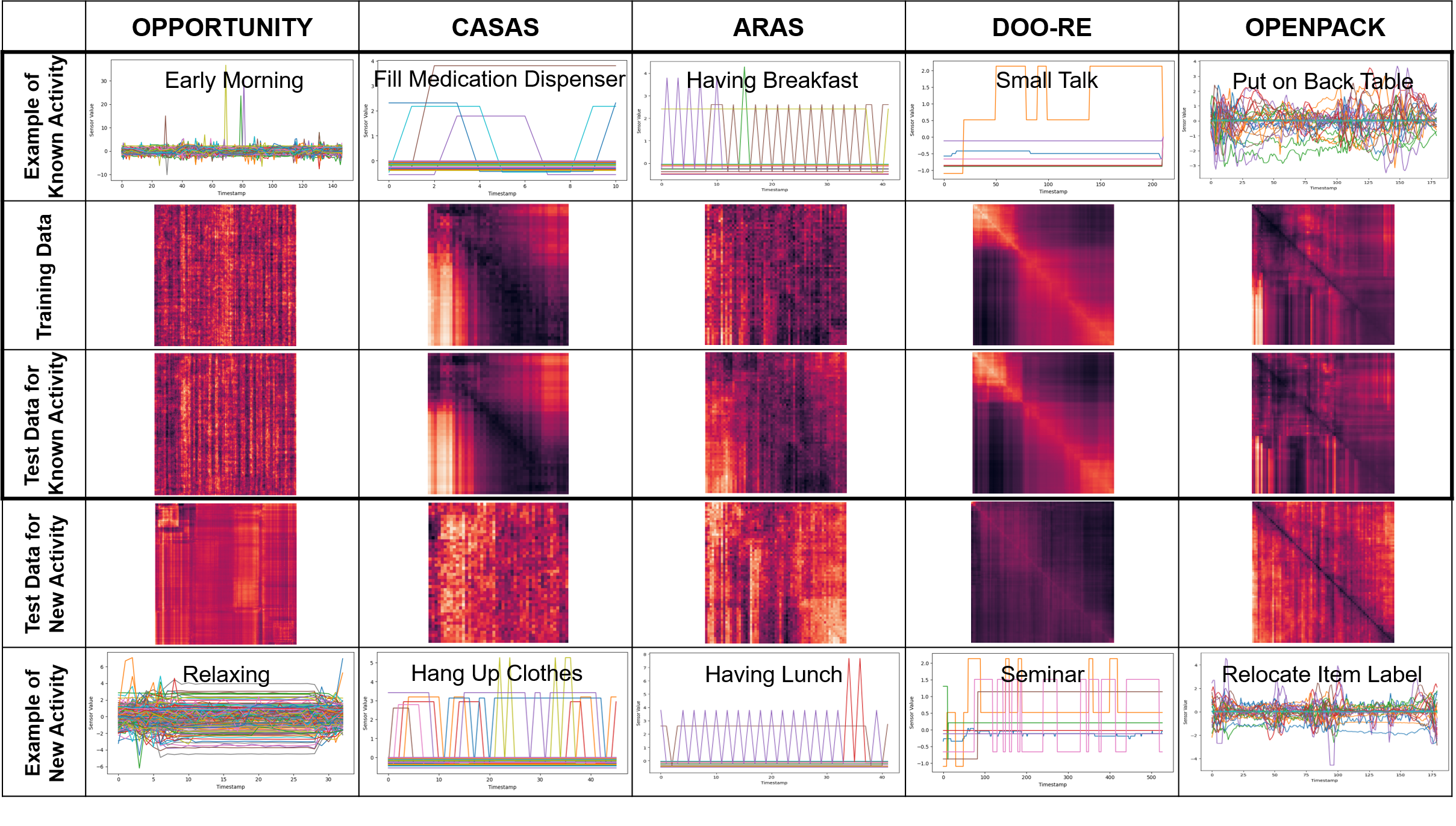}
    \caption{The attention maps visualized by \texttt{CLAN} highlight the representational differences between known and new activities. 
    Brighter colors indicate higher attention weights.}
    \label{fig 9: attentionmaps}
\end{figure}

\subsection{Qualitative Analysis}\label{visualization}

\subsubsection{Representations for Each Dataset from \texttt{CLAN}.}  
To qualitatively evaluate \texttt{CLAN}'s capability to generate discriminative representations and its explainability in practical applications \cite{transformer,interpretability}, Fig.~\ref{fig 9: attentionmaps} presents attention maps from \texttt{CLAN}'s Transformer encoder, visually depicting known and novel activity data across datasets.  

(1) For all datasets, the attention map patterns of training and test data for known activities exhibit remarkable similarities, demonstrating that \texttt{CLAN} effectively learns invariant representations of known activities.  
(2) The representations of test data for new activities differ significantly from those of the known activities, highlighting \texttt{CLAN}'s ability to detect representational novelties as new activities.  
Even for similar activities, such as \textit{Having Breakfast} (Known) and \textit{Having Lunch} (New) in \textit{ARAS}, \texttt{CLAN} derives distinct key patterns for each.  
(3) The visualization further reveals the core features of each activity.  
For example, in \textit{DOO-RE}, early and late-stage patterns in attention maps are crucial for the known activity \textit{Small Talk}, whereas in \textit{CASAS}, patterns occurring from early to mid-late stages are significant for the known activity \textit{Filling the Medication Dispenser}. 
This visualization method provides valuable insights into interpreting both known and new activities in practical applications.

\subsubsection{Separation Ability.}  

\begin{figure}[t]
    \centering
    \includegraphics[width= 0.9\linewidth]{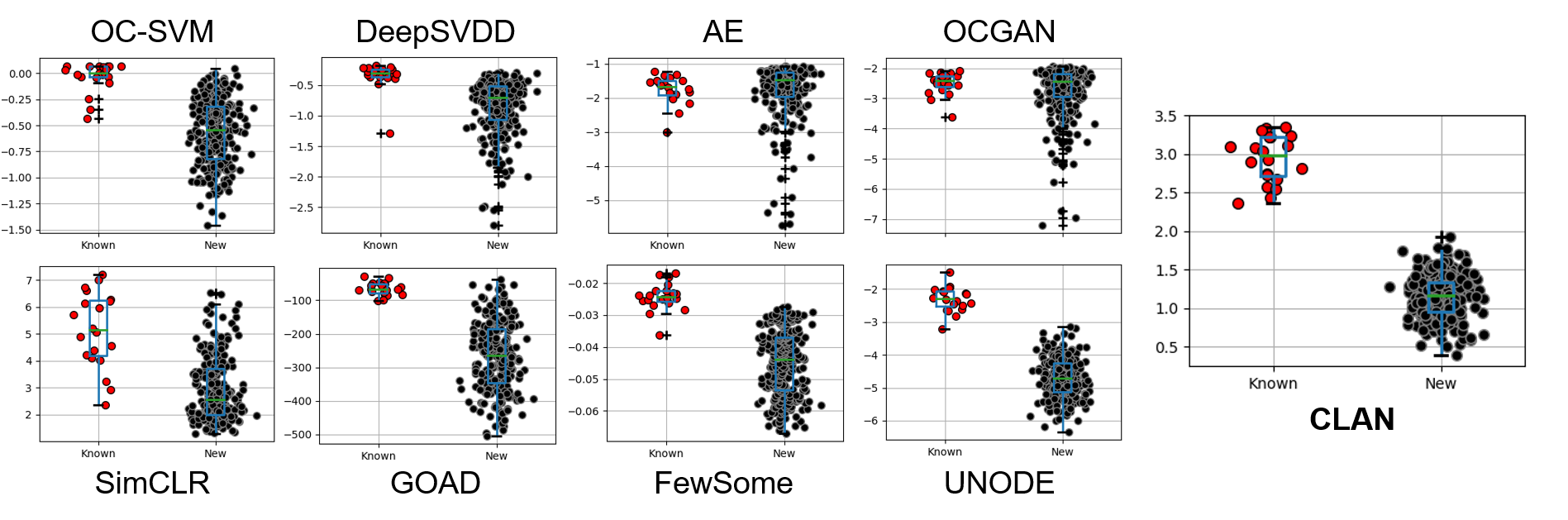}
    \caption{Scatter plots illustrating the distribution of new activity detection scores for each method when \textit{Play a Game of Checkers} (6) in \textit{CASAS} is set as the known activity.}
    \label{fig 11: plotgrplot}
\end{figure}

To qualitatively evaluate the distinctiveness of representations across novelty detection approaches, we employ scatter plots supplemented by box plots, as illustrated in Fig.~\ref{fig 11: plotgrplot}.  
Since different approaches utilize varying representation learning and scoring techniques, novelty detection scores serve as the only common metric for comparison.  
We use scatter plots based on the novelty detection scores to fairly assess the discriminative capability of each method.  

(1) Notably, \texttt{CLAN}'s new activity scores (black dots) form a tighter cluster compared to other baselines, demonstrating that \texttt{CLAN} effectively extracts high-quality invariant representations for known activities.  
(2) The box plots superimposed on scatter plots reveal minimal overlap between the score distributions of known and new activities in \texttt{CLAN}, highlighting the effectiveness of its discriminative representation-based detection mechanism.  
These visual results further confirm the robustness of \texttt{CLAN} in capturing new activities.  

%% file: content/6.discussion.tex
\section{Discussion}

\begin{figure}[b]
  \begin{minipage}[c]{.5\linewidth}
    \centering
    \includegraphics[width=\linewidth]{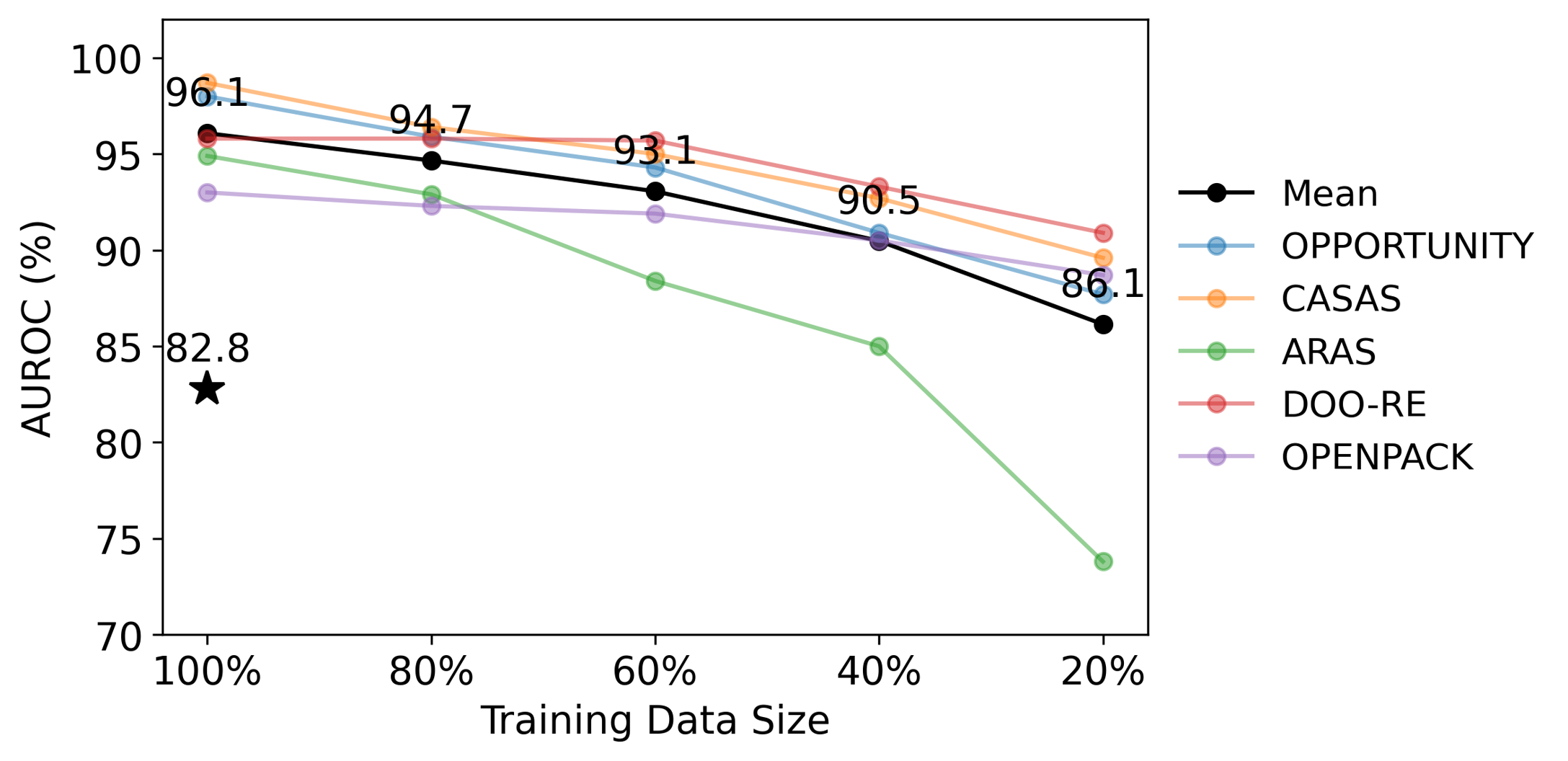}
    \captionof{figure}{AUROC(\%) according to training data size.
    The black star is the mean of the top baseline (FewSome) performance.}
    \label{fig 12: data_size}
  \end{minipage}\hspace{1em}
  \begin{minipage}[c]{.4\linewidth}
    \centering
    \begin{tabular}{c|cc}
    \hline
    {Dataset} & {FewSome} & {\texttt{CLAN}} \\
    \hline
    {OPPORTUNITY} & 24.558 & 3.925 \\
    {CASAS}       & 19.239  & 2.712 \\
    {ARAS}        & 18.178	 & 2.988 \\
    {DOO-RE}      & 21.208  & 2.729 \\
    {OPENPACK}    & 17.996 & 2.976 \\\hline
    {Average}     & 20.236 & 3.066 \\
    \hline
    \end{tabular}    
    \captionof{table}{New activity detection inference time (\textit{ms}) for all datasets.}
    \label{Tab7: Time}
  \end{minipage}
\end{figure}

\noindent\textbf{Performance Robustness to Training Data Size.}
Due to the data-scarce scenario being a significant challenge in HAR \cite{data_shift, TF_wearable}, we evaluate the impact of varying training data sizes for known activities on \texttt{CLAN}, as shown in Fig.~\ref{fig 12: data_size}.  
The evaluation starts with the full training dataset and gradually reduces the data size in 20\% increments, down to 20\% of the original size.

Even with this limited data, \texttt{CLAN} achieves remarkable results, scoring 87.7\% in \textit{OPPORTUNITY}, 89.6\% in \textit{CASAS}, 73.8\% in \textit{ARAS}, 90.9\% in \textit{DOO-RE}, and 88.7\% in \textit{OPENPACK}.  
Compared to the baseline performances in Table~\ref{tab 2: one-class}, \texttt{CLAN} surpasses most baselines even when trained with only 20\% of the data.  
These findings validate \texttt{CLAN}'s practicality in real-world scenarios, demonstrating its ability to accurately detect new activities even with limited known activity data.

\noindent \textbf{Inference Time.}
A comparative analysis is conducted to evaluate the inference speed of \texttt{CLAN} relative to \textit{FewSome}, the best-performing baseline known for its efficiency in real-world applications.  
\textit{Inference time} is defined as the average computational duration (\textit{ms}) required to extract features and compute a new activity score for a single sample.  
To ensure precise measurement of computation time on both GPU and CPU, \texttt{torch.cuda.synchronize}\footnote{https://pytorch.org/docs/stable/generated/torch.cuda.synchronize.html} is utilized.  

As summarized in Table~\ref{Tab7: Time}, \texttt{CLAN} achieves a substantial speed advantage, running $6.05 \times$ to $7.77 \times$ faster than the baseline method.  
Across all datasets, \texttt{CLAN} not only maintains superior accuracy but also demonstrates higher inference efficiency.  
This improvement in computational speed is attributed to \texttt{CLAN}’s streamlined new activity detection mechanism, which optimizes similarity and auxiliary classification scoring.  
These results strongly support the feasibility of integrating \texttt{CLAN} into real-world applications.

\begin{figure}[t]
    \centering
    \includegraphics[width=0.8\linewidth]{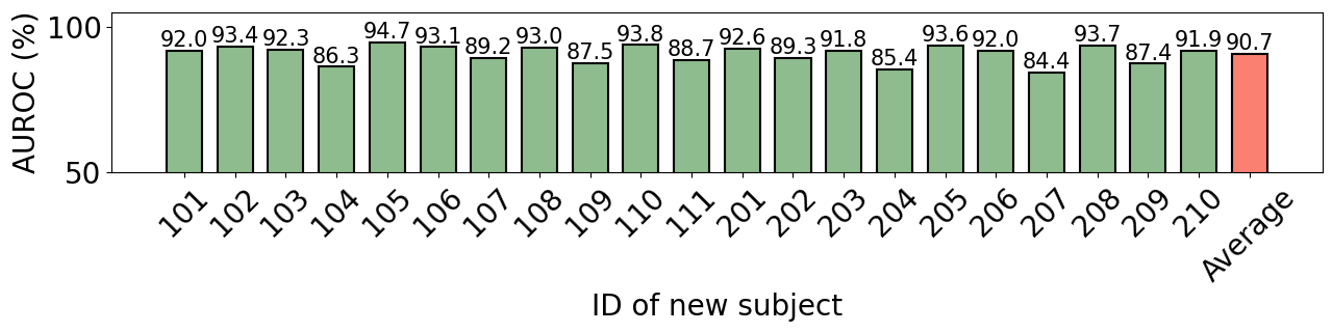}
    \caption{AUROC(\%) for new activity detection in new \textit{unseen} subjects on \textit{OPENPACK}. 
    The x-axis represents the IDs of each new user for whom no information is provided during the training phase.
    }
    \label{shift test}
\end{figure}

\noindent \textbf{Performance in the Cross-domain Setting.}
To evaluate \texttt{CLAN}'s ability to detect new activities in new subjects under data distribution shifts—an essential and recent research challenge in the HAR domain \cite{data_shift}—we conduct experiments on new users using \textit{OPENPACK}.  
In this scenario, both the label spaces and distributions differ, such that $X_{s} = X_{t}$, but $Y_{s} \neq Y_{t}$ and $p(x)_{s} \neq p(x)_{t}$.  
The experiments follow a leave-one-person-out cross-validation (LOOCV) setting, where all known and new activity data for the new users are excluded during training.  

As shown in Fig.~\ref{shift test}, performance varies across new users, but the overall average reduction is only about 2\%, while still outperforming existing baselines.  
Thus, \texttt{CLAN} still maintains robustness in detecting new activities even under data distribution shifts with new users, demonstrating its adaptability to real-world scenarios.  
To further enhance performance in cross-domain settings, we will integrate \texttt{CLAN} with LLM-enriched techniques and domain adaptation methods \cite{cross-domain, UniMTS}, improving its scalability across diverse environments.

%% file: content/7.conclusion.tex
\section{Conclusion and Future work}

In this paper, we propose \texttt{CLAN}, a two-tower model 
that leverages \textbf{C}ontrastive \textbf{L}earning with diverse data \textbf{A}ugmentation for \textbf{N}ew activity detection in the open-world.
\texttt{CLAN} enables new activity detection using only known activities and is tailored to the unique properties of each sensor-based environment.
It leverages self-supervised learning techniques to construct discriminative representations by comparing various types of negatives across both the time and frequency domains.
Extensive experiments on real human activity datasets demonstrate that \texttt{CLAN} outperforms existing novelty detection methods and validates its effectiveness in real-world scenarios.

A potential limitation of \texttt{CLAN}, similar to other self-supervised approaches, is its reliance on effective data augmentation methods.
Currently, \texttt{CLAN} utilizes commonly used augmentation techniques, which may constrain its performance and may not be fully optimized for new activity detection.
To address this limitation, we plan to develop a novel data augmentation technique that generates multi-faceted and optimally tailored samples for each dataset by leveraging recent advancements in generative AI technologies \cite{genai}.

%% file: content/appendix.tex
\appendix

\section{Details of Datasets} \label{A. dataset_details}
Table~\ref{Tab 8: activity types} outlines the activities present in each dataset, while Table~\ref{Tab 9: sensor types} details the sensor types used in each dataset.

\begin{table*}[h]
\centering
\caption{A description of the activities in each dataset. Numbers in parentheses indicate the assigned activity numbers.
}
\scalebox{0.9}{
\begin{tabular}{c | p{4.9in}}\hline\label{Tab 8: activity types}
\textbf{Datasets}&\textbf{Activity Types}\\\hline
OPPORTUNITY
& \textit{Relaxing(1)}, \textit{Coffee time(2)}, \textit{Early morning(3)}, \textit{Clean up(4)}, \textit{Sandwich time(5)}
\\\hline
CASAS
&   \textit{Fill medication dispenser(1)},
    \textit{Hang up clothes in the hallway closet(2)},
    \textit{Move the couch and coffee table(3)},
    \textit{Sit on the couch and read a magazine(4)},
    \textit{Sweep the kitchen floor(5)},
    \textit{Play a game of checkers(6)},
    \textit{Set out ingredients(7)},
    \textit{Set dining room table(8)},
    \textit{Read a magazine(9)},
    \textit{Simulate paying an electric bill(10)},
    \textit{Gather food for a picnic(11)},
    \textit{Retrieve dishes from a kitchen cabinet(12)},
    \textit{Pack supplies in the picnic basket(13)},
    \textit{Pack food in the picnic basket(14)}
\\\hline
ARAS
& \textit{Going Out(2)}, 
    \textit{Preparing Breakfast(3)}, 
    \textit{Having Breakfast(4)}, 
    \textit{Preparing Lunch(5)}, 
    \textit{Having Lunch(6)}, 
    \textit{Preparing Dinner(7)}, 
    \textit{Having Dinner(8)}, 
    \textit{Washing Dishes(9)}, 
    \textit{Having Snack(10)}, 
    \textit{Sleeping(11)}, 
    \textit{Watching TV(12)}, 
    \textit{Studying(13)}, 
    \textit{Having Shower(14)}, 
    \textit{Toileting(15)}, 
    \textit{Napping(16)}, 
    \textit{Other(1)}
\\
\hline
DOO-RE
& \textit{Small talk(1)}, \textit{Studying together(2)}, \textit{Technical discussion(3)},  \textit{Seminar(4)}
\\
\hline
OPENPACK
& 
\textit{Picking(1)}, 
\textit{Relocate item label(2)}, 
\textit{Assemble box(3)}, 	
\textit{Insert items(4)},
\textit{Close box(5)}, 
\textit{Attach box label(6)}, 
\textit{Scan label(7)}, 
\textit{Attach shipping label(8)}, 	
\textit{Put on back table(9)}, 
\textit{Fill out order(10)}
\\
\hline
\end{tabular}}
\end{table*}

\begin{table*} [h]
\centering
\caption{A description of the sensors used in each dataset.}
\scalebox{0.9}{
\begin{tabular}{c|p{4.9in}}\hline\label{Tab 9: sensor types}
\textbf{Datasets}&\textbf{Sensor Types}\\\hline
OPPORTUNITY
& \textit{242 sensors including Spoon accelerometer (accX, gyroX, gyroY), Sugar jar accelerometer (accX, gyroX, gyroY), Dishwater reed switch, Fridge reed switch, Left shoe inertial measurement, User location, etc.
For more detailed sensor information, see \url{https://archive.ics.uci.edu/ml/datasets/opportunity+activity+recognition}.
}
\\\hline
CASAS
& \textit{Motion (M01...M51)},
\textit{Item (I01..I08)},
\textit{Cabinet (D01..D12)}
\\\hline
ARAS
& \textit{Wardrobe photocell},
\textit{Convertible couch photocell},
\textit{TV receiver IR},  
\textit{Couch force }, 
\textit{Couch force\_2},
\textit{Chair distance},
\textit{Chair distance\_2},
\textit{Fridge photocell},
\textit{Kitchen drawer photocell},
\textit{Wardrobe photocell},
\textit{Bathroom cabinet photocell},
\textit{House door contact },
\textit{Bathroom door contact},
\textit{Shower cabinet door contact },
\textit{Hall sonar distance},
\textit{Kitchen sonar distance},
\textit{Tap distance},
\textit{Water closet distance},
\textit{Kitchen temperature},
\textit{Bed force }
\\
\hline
DOO-RE
& \textit{Seat occupation}, 
\textit{Sound level},
\textit{Brightness level},
\textit{Light status}, 
\textit{Existence status}, 
\textit{Projector status}, 
\textit{Presenter detection status}
\\
\hline
OPENPACK
&
\textit{IMUs (each including Acc, Gyro, Quaternion) for right wrist, left wrist, right upper arm and left upper arm},
\textit{E4 sensors (each including Acc, BVP, EDA, Temperature) for right wrist and left wrist},
\textit{IoT device sensors for a handheld scanner and a label-printer}\\\hline
\end{tabular}
}
\vspace{-15pt}
\end{table*}